\documentclass[11pt]{article}

\usepackage[final]{acl}
\usepackage{times}
\usepackage{latexsym}
\usepackage[T1]{fontenc}
\usepackage[utf8]{inputenc}
\usepackage{microtype}

\makeatletter
\renewcommand{\verbatim@font}{\scriptsize\ttfamily}
\makeatother
\usepackage{graphicx}

\usepackage{booktabs}
\usepackage{amsfonts}
\usepackage{nicefrac}
\usepackage{multirow}
\usepackage{array}

\title{SocialMemBench: Are AI Memory Systems Ready\\for Social Group Settings?}

\author{Olukunle Owolabi \\
  Meta\thanks{Disclaimer: While Author is currently employed by Meta, this work was performed in Personal capacity outside of Meta Employment. Meta does not endorse the content.} \\
  \texttt{olukunle.owolabi@alumni.tufts.edu}}

\begin{document}

\maketitle

\begin{abstract}
Memory systems for AI assistants were built for single-user dialogue, a design mismatched to multi-party social group settings. This matters for the social assistants being built today: group-acting agents in chat platforms, and proactive personal assistants whose model of a user must include their social context. Existing memory benchmarks evaluate dyadic or workplace dialogue; none targets social groups, where memory must anchor facts in shared history rather than professional roles, separate group norms from individual exceptions, and attribute correctly even after a member leaves.

We introduce SocialMemBench, a benchmark of human-verified synthetic social group networks across five archetypes and three size tiers (4--30 members), with 430 personas and 1,031 QA pairs across nine question categories. Our evaluation exposes five failure modes, framed as testable hypotheses, and we provide evidence on two through the research probes Subject-Mem and SMG.

Our results support three connected takeaways. First, social-group memory must preserve distinctions that ordinary fact recall misses, such as who dissented from a decision and what one member knows about another. Second, across 43 networks the four evaluated frameworks (Mem0, LangMem, Graphiti, Cognee) score 0.12--0.18, well below uncompressed retrieval (0.346) and full context (0.370) with the same answerer, and changing the answerer or retrieval budget does not close the gap. Third, explicitly representing social information mitigates specific failure modes but leaves others open: Subject-Mem targets single-stream conflation, lifting attribution from 0.25--0.46 to 0.78, and SMG targets norm-individual conflation, lifting group-decision accuracy to 0.70, while the other three modes remain open. We release the dataset, the evaluation pipeline with both research probes, and the audit viewer as a testbed for social-memory research.\footnote{\label{fn:urls}Dataset: \url{https://huggingface.co/datasets/qunlexie/SocialMemBench} (CC BY 4.0). Pipeline and audit viewer: \url{https://github.com/Qunlexie/SocialMemBench} (MIT).}

\end{abstract}

\section{Introduction}

AI assistants are increasingly deployed alongside multi-person conversations in two distinct modes. Group-acting agents embedded in social and messaging platforms take actions on behalf of the group;\footnote{Examples include Meta AI in WhatsApp group chats, Salesforce Agentforce in Slack channels, and Apple Intelligence's on-device personal-context retrieval \citep{gunter2024apple}.} proactive personal-assistant agents \citep{generativeagents2023, zhong2024memorybank, memgpt2023} build holistic models of their users, for which social context is a vital input. Both classes share a single dependency: a memory layer that retains social facts across sessions and attributes them to the right person.

Current memory frameworks were designed for a single user \citep{zhang2024survey}. Mem0 \citep{mem0} keys facts to a single user; LangMem \citep{langmem} namespaces memory under that user; Graphiti \citep{graphiti} merges same-named entities into canonical nodes; HippoRAG \citep{hipporag2024} and Generative Agents \citep{generativeagents2023} both treat memory as a single stream owned by one observer. These designs are correct for the dyadic chatbot setting they were built for, but are mismatched to social groups, where attribution must rest on shared history and informal norms rather than the professional roles that partially scaffold workplace dialogue. The architectural problems are concrete: a statement one member makes about another is stored against the speaker, not the subject; group consensus overwrites individual dissent rather than preserving it as an exception; a member's preferences are versioned by recency rather than by who held what when. Speaker-aware reasoning failures in multi-party dialogue are documented at the LLM level \citep{fantom2023, sclar2024minding}, and attribution remains a challenge in workplace conversations even with the full conversation in context \citep{evermembench2026}. The social setting, where role-scaffolding is absent and attribution must come from the structure of the conversation itself, has not been benchmarked at the memory-architecture level.

This gap is not directly addressed by existing memory benchmarks. Prior multi-session memory benchmarks operate in dyadic chat or workplace settings: LongMemEval \citep{longmemeval2025} evaluates user-to-assistant memory across five ability dimensions but is dyadic, and EverMemBench \citep{evermembench2026} tests multi-party attribution in a workplace setting where professional roles partially scaffold attribution. Social group dialogue has no such scaffolding; memory must anchor facts in shared history rather than professional roles, separate group norms from individual exceptions, and correctly attribute even after a member leaves the group.

We introduce SocialMemBench to address this gap: human-verified synthetic social group networks across five archetypes (close friends, family, recreational, interest community, acquaintance network) and three group-size tiers (4--30 members), with 430 personas, 348 sessions, 7,355 conversation turns, and 1,031 QA pairs across nine social-memory question categories.

Our results support three connected takeaways. First, social-group memory must preserve distinctions that ordinary fact recall misses: a decision versus who dissented, a preference versus what another member knows about it, and preferences that change or members who leave (Section~\ref{sec:query-taxonomy}). Second, the evaluated frameworks score 0.12--0.18, well below references that preserve the original conversation (0.346 for uncompressed retrieval, 0.370 for full context, same answerer), and changing the answerer or retrieval budget does not close the gap (Table~\ref{tab:main-results}). Third, explicitly representing social information mitigates specific failure modes but leaves others open: subject indexing (Subject-Mem) targets single-stream conflation, lifting attribution from 0.25--0.46 to 0.78, and a dissent-aware schema (SMG) targets norm-individual conflation, lifting group decisions to 0.70, while the other three modes remain open (Section~\ref{sec:failure-modes}).

Our contributions are: (i) SocialMemBench, 43 human-verified synthetic social group networks with 430 personas and 1,031 QA pairs across nine categories; (ii) an evaluation of four open-source memory frameworks, two research probes, and two conversation-preserving references; (iii) five architectural failure modes grounded in the results, two targeted by our probes; and (iv) an open evaluation pipeline and audit viewer for human review.\textsuperscript{\ref{fn:urls}}

\section{Related Work}

\paragraph{Memory and Long-Context Benchmarks.}
The challenge of preserving facts across extended conversations has been studied at progressively larger scale. \citet{msc2022} introduced a canonical dataset of persona-grounded dyadic conversations spanning up to five sessions, establishing multi-session coherence as an evaluation target. \citet{locomo2024} scaled this to very long-term dyadic dialogues of about 300 turns over up to 35 sessions, with persona and temporal-event-graph grounding, and added multi-hop and temporal QA. \citet{longmemeval2025} formalized user-to-assistant memory as five ability dimensions (single-session extraction, cross-session reasoning, temporal reasoning, knowledge updates, and abstention), finding 30--60\% accuracy drops when chat histories span multiple sessions. \citet{evermembench2026} breaks the dyadic ceiling: it tests multi-party workplace conversations and reports that multi-hop accuracy falls to 26.5\% for Gemini 3 Flash given the full conversation, against 88.4\% when the gold evidence spans are supplied. A separate body of work evaluates whether LLMs can reason over long documents within a single context window. SCROLLS \citep{scrolls2022} established a unified suite of long-context tasks spanning summarization, question answering, and natural language inference, and subsequent benchmarks have progressively extended context lengths while showing that claimed context windows overstate real capability.

\paragraph{Memory Frameworks.}
Several open-source memory frameworks for LLM agents have emerged. Mem0 \citep{mem0} extracts atomic facts with semantic deduplication; Graphiti \citep{graphiti} builds a temporally-aware entity-relationship graph with validity windows and provenance; LangMem \citep{langmem} uses namespace-keyed storage with reflection-driven consolidation; Cognee \citep{cognee} combines graph and vector storage with LLM-extracted facts. MemGPT \citep{memgpt2023} introduced tiered virtual memory with LLM-driven paging, and Generative Agents \citep{generativeagents2023} layered recency, importance, and relevance scoring over a natural-language memory stream. HippoRAG \citep{hipporag2024} applies graph-structured retrieval with personalized PageRank over LLM-extracted entities.

\paragraph{Personalization and Social Structure.}
Personalization research addresses individual user modeling. LaMP \citep{lamp2024} established the canonical benchmark for single-user personalization, and subsequent retrieval- and parameter-based work has used it as the evaluation target. PersonaChat \citep{personachat2018} models persona as short in-context text fragments rather than maintained cross-session state. Social NLP has studied structure within conversations. DialogRE \citep{dialogre2020} extracted typed relations between speakers within single dialogues; graph-based emotion recognition models encode speaker structure per session. \citet{gandhi2023bigtom} introduce BigToM, a model-generated theory-of-mind benchmark, and find that GPT-4's inferences follow human patterns but less reliably, while other models struggle. FANToM \citep{fantom2023} tests theory-of-mind reasoning in multi-party conversations where information asymmetry arises from characters with unequal access to prior exchanges.

Table~\ref{tab:benchmark-comparison} compares these benchmarks across architectural dimensions.

\begin{table*}[t!]
  \centering
  \small
  \begin{tabular}{lcccc}
    \toprule
    \textbf{Benchmark} & \textbf{Multi-party} & \textbf{External memory} & \textbf{Social setting} & \textbf{Member departure} \\
    \midrule
    MSC \citep{msc2022} & -- & -- & -- & -- \\
    LoCoMo \citep{locomo2024} & -- & -- & -- & -- \\
    LongMemEval \citep{longmemeval2025} & -- & \checkmark & -- & -- \\
    LaMP \citep{lamp2024} & -- & -- & -- & -- \\
    FANToM \citep{fantom2023} & \checkmark & -- & -- & -- \\
    EverMemBench \citep{evermembench2026} & \checkmark & \checkmark & -- & -- \\
    \textbf{SocialMemBench (ours)} & \checkmark & \checkmark & \checkmark & \checkmark \\
    \bottomrule
  \end{tabular}
  \caption{Benchmark comparison across architectural dimensions. ``Multi-party'' = $\geq$3 co-present speakers per session. ``External memory'' = tests a write-index-retrieve layer across discrete sessions where the original conversation is not re-supplied at query time (vs. LLM reasoning over full context). ``Social setting'' = informal social group (friends, family, recreational, interest community), not workplace, dyadic strangers, or single-user-to-assistant. ``Member departure'' = corpus contains members who leave mid-corpus with explicit ground truth for pre-departure state recall.}
  \label{tab:benchmark-comparison}
\end{table*}
\section{Benchmark Design}

\subsection{Task Definition}
\label{sec:task-def}

Let $G = (P, E, S)$ be a social group with personas $P$, pairwise relationship edges $E$, and a session sequence $S = (s_1, \ldots, s_T)$. A memory system $M$ ingests sessions sequentially, extracting facts and constructing a memory state after each one. After all sessions are ingested, $M$ answers a question $q$ about a specific persona or group interaction using only that memory state; the original conversation is not re-supplied as context. SocialMemBench measures how accurately $M$ can answer $q$ under this setup. A full-context reference is reported separately by giving an LLM answerer the full conversation as context rather than a memory state. The full-context setting is a reference point rather than a system under test. Persistent memory lets an assistant carry information across sessions without supplying the whole history on every query, and larger context windows do not remove the need to evaluate what survives memory construction. Prior work reports cost and latency savings for retrieval and conversational memory relative to long-context inference \citep{li2024rag, mem0}; we do not measure efficiency at our history lengths. Our question is the complementary one: whether the facts a social group needs survive when the conversation is compressed into stored memory.

\subsection{Query Taxonomy}
\label{sec:query-taxonomy}

The benchmark covers nine proposed question categories, each targeting a distinct capability that a social-group AI needs. Table~\ref{tab:query-types} lists the categories with examples and the key difficulty driver for each. Each category contains many questions across many networks; the labels are category identifiers, not question numbers.

The categories combine established memory-ability dimensions with capabilities specific to multi-party groups. SR, MA, TS, and DM instantiate the dyadic memory abilities of LongMemEval \citep{longmemeval2025} (extraction, multi-session and temporal reasoning, knowledge updates, abstention). AP, TM, GD, NI, and RE are social extensions that dyadic and workplace taxonomies omit: AP targets the multi-party attribution failure reported by \citet{evermembench2026}, TM follows information-asymmetry theory-of-mind \citep{fantom2023}, NI builds on non-monotonic social-norm reasoning \citep{normbank2023}, and GD and RE capture consensus and relationship history that arise only in groups. We do not claim to enumerate every memory capability, only those a memory layer must support in social settings.

Three categories test attribution specifically: attribution probes (AP, direct speaker confusion), theory-of-mind (TM, who knows what about whom), and norm-vs-individual (NI, does a group-level fact apply to this specific person).

\begin{table*}[t!]
  \centering
  \small
  \begin{tabular}{p{0.5cm} p{2.4cm} p{7.9cm} p{3.5cm}}
    \toprule
    & \textbf{Category} & \textbf{Example question} & \textbf{Difficulty driver} \\
    \midrule
    SR & Single-contact recall & What does Hiro's behavior in reunion food planning suggest about his eating pattern, and how is it visible without him stating it directly? \textcolor{blue}{(close friends \textperiodcentered{} OE)} & Implicit signal; multi-turn synthesis \\
    GD & Group decision recall & When Bex closes the headcount for the rooftop component, what did Aanya's response signal about her actual position? \textcolor{blue}{(close friends \textperiodcentered{} MC)} & Consensus decision plus dissenter identity \\
    MA & Multi-contact aggregation & All five Maple Street members shared their views on the residents' parking scheme. What was each person's position and reasoning? \textcolor{gray}{(acquaintance \textperiodcentered{} OE)} & Per-persona enumeration; some preferences implicit \\
    AP & Attribution probe & Who expressed the view that the value of a photography session lies in intentionality rather than the number of exposures? \textcolor{purple}{(interest community \textperiodcentered{} MC)} & Two plausible speakers; attribution by meaning \\
    TM & Theory of mind & What is Priya's concern about running on uneven surfaces, and what in Fatima's response shows she already knew about this? \textcolor{orange}{(recreational \textperiodcentered{} OE)} & Cross-persona knowledge revealed by behavior \\
    NI & Norm vs. individual & The group has a norm of raising issues in the chat before contacting the council. Does that norm apply to everyone? \textcolor{gray}{(acquaintance \textperiodcentered{} MC)} & Norm-exception distinction; implicit dissent \\
    RE & Relational edge & What does the conversation reveal about Bram and Lina's relationship, particularly through the First Watt loan exchange? \textcolor{purple}{(interest community \textperiodcentered{} OE)} & History embedded in passing references \\
    TS & Temporal shift & What in the conversation signals that Priya's relationship to running and team sport has shifted, and what appeared to drive it? \textcolor{blue}{(close friends \textperiodcentered{} OE)} & Old state + trigger + new state; cross-session \\
    DM & Departed member & Based on what the group knew about Bex before she left, what was her relationship to cycling? \textcolor{blue}{(close friends \textperiodcentered{} OE)} & Stale-state recall post-departure \\
    \bottomrule
  \end{tabular}
  \caption{SocialMemBench query categories. Each targets a distinct memory capability. Category codes are used as compact column headers in results tables. Inline tags show the archetype the example came from and the answer format: ``OE'' = open-ended (LLM-judge scored 0--1), ``MC'' = multiple-choice (scored by exact letter match).}
  \label{tab:query-types}
\end{table*}
\subsection{Data Generation Pipeline}
\label{sec:data-gen}

The corpus is synthetically generated. Synthetic generation is not a shortcut but a methodological choice: it lets us plant ground-truth challenges at known positions, enforce per-persona voice fingerprints verified by a blind critic, and control implicitness levels across nine question categories. BABILong~\citep{kuratov2024babilong}, SODA~\citep{kim2023soda}, RULER~\citep{hsieh2024ruler}, and Self-Instruct~\citep{wang2023selfinstruct} use synthetic generation for the same reason: to guarantee evaluation properties that naturally-occurring data cannot provide. Concretely, the annotations it requires, verbatim evidence spans at known (speaker, session, turn) positions, subject attribution for third-party mentions, and verifiable pre-departure states, do not exist in real chat corpora.

Generation runs in three stages, each an LLM pass (Claude Sonnet 4.5) whose output feeds the next and is verified before proceeding (Figure~\ref{fig:pipeline}); Appendix~\ref{app:worked-example} traces one planted fact through all three.

\paragraph{Stage 1: Persona Network.} Each network specifies $n$ personas with profiles, communication styles, preference histories, and pairwise relationship edges, under structural constraints verified before proceeding: at least one contested group norm (ground truth for NI), one preference-history event per persona (ground truth for TS), and distinct communication profiles. Networks span three size tiers (Table~\ref{tab:dataset-stats}).

\paragraph{Stage 2: Conversation Corpus.} For each network we generate $S$ sessions, each a self-contained WhatsApp-style thread rendered with session and speaker headers. Each session embeds 2--4 planted challenges: ground-truth facts expressed implicitly through behavior, complaint, or deflection rather than stated directly, each recorded with its subject, speaker, fact, and a verbatim evidence span. Challenge types cover the query taxonomy (Table~\ref{tab:challenge-types}, appendix).

\paragraph{Stage 3: QA Generation and Verification.} Each planted challenge seeds one or more questions. A pre-generation inference gate rejects any question answerable by keyword search and enforces that difficulty labels match required reasoning depth. After generation, a blind-critic pass (an LLM evaluator shown only the plain-text corpus, with no access to planted challenge metadata) independently answers every question and scores it against the ground truth. A network is accepted only when the blind-critic mean score $\geq 0.70$ (the rubric boundary between an answer recoverable from the text and one only partially recoverable) and all evidence spans verify verbatim; admission is insensitive to this threshold (Table~\ref{tab:threshold-sensitivity}). Every QA pair was additionally reviewed by a human using the browser-based data audit viewer (Appendix~\ref{app:viewer}), checking plausibility, attribution correctness, and answer completeness; pairs failing this review were revised or removed (full criteria in Appendix~\ref{app:data-construction}).

\subsection{Dataset}
\label{sec:dataset-stats}

The corpus is 348 sessions of naturalistic group dialogue across 43 networks (430 personas, 660 pairwise relationship edges): 7,355 conversation turns averaging 21.1 turns per session, 292,725 tokens total under the GPT-4o-mini tokenizer. Each session is a self-contained chat exchange generated to embed 2--4 planted challenges; each challenge records the subject persona, the speaker, a verbatim evidence span, and an implicitness level. Of the 1,091 planted challenges, 89\% carry implicitness level 2 or 3, meaning the fact is expressed through behavior, complaint, or deflection rather than stated directly. Every question in the benchmark has a traceable ground-truth anchor in the text: a specific turn where the relevant fact is revealed. Networks span five group types (close friends, family, recreational, interest community, acquaintance network), ranging from strong ties (close friends, family) to weak ties (acquaintance networks), covering the principal social contexts in which group-deployed AI assistants operate. Attribution patterns, norm structures, and relationship histories differ meaningfully across group types; a system that handles only close-friend networks has not solved the general problem.

Questions cover all nine categories listed in Table~\ref{tab:query-types}; per-category counts are reported in Appendix~\ref{app:data-construction} (Table~\ref{tab:category-counts}). Blind-critic mean scores range 0.93--0.98 across categories (overall mean 0.95), confirming every question is answerable from the corpus by a reader with full access. A system that stored only explicitly stated preferences would score near zero on the 83\% of questions backed by implicit challenges.

\begin{table*}[t!]
  \centering
  \small
  \begin{tabular}{lcccccc}
    \toprule
    \textbf{Tier} & \textbf{Nets} & \textbf{Members} & \textbf{Personas} & \textbf{Sessions} & \textbf{Turns} & \textbf{QA} \\
    \midrule
    Small & 19 & 4--6 & 95 & 122 & 2,183 & 264 \\
    Medium & 14 & 8--10 & 130 & 124 & 2,562 & 346 \\
    Large & 10 & 15--30 & 205 & 102 & 2,610 & 421 \\
    \midrule
    \textbf{Total} & \textbf{43} & & \textbf{430} & \textbf{348} & \textbf{7,355} & \textbf{1,031} \\
    \bottomrule
  \end{tabular}
  \caption{Corpus statistics by network-size tier. All 43 networks passed structural verification, blind-critic QC, and were evaluated across all 8 conditions (19 small including 4 with member-departure scenarios, 14 medium, 10 large).}
  \label{tab:dataset-stats}
\end{table*}
\section{Evaluation Framework}

\subsection{Design}
\label{sec:conditions}

We evaluate four open-source memory frameworks representative of current architectural approaches: \textsc{graphiti} (Zep AI), \textsc{langmem}, \textsc{mem0}, and \textsc{cognee}. We also evaluate two research probes built for this benchmark, each a minimal architectural variant of a standard framework. \textsc{subject-mem} (Subject-Indexed Memory) indexes facts by the subject persona rather than the speaker. \textsc{smg} (Social Memory Graph) uses a structured graph with dissent-aware and relational edges instead of flat embeddings.

Two reference baselines hold the answerer constant to isolate memory-layer effects. The \textit{full-context baseline} (\textsc{llm-mini}, GPT-4o-mini) receives the complete corpus with no memory or retrieval step; the \textit{uncompressed retrieval baseline} stores every turn in a vector store and retrieves by cosine similarity, without extraction or summarization.

The primary evaluation uses GPT-4o-mini as the answerer and top-$k$=10 retrieval where retrieval is used. The non-reasoning answerer is a deliberate design choice: a strong reasoning model with extensive chain-of-thought can compensate for incomplete or misattributed retrievals through inference, masking the architectural failure modes the benchmark is built to expose. The architectural claims we make in Sections~\ref{sec:results}--\ref{sec:failure-modes} describe information loss prior to retrieval and are model-agnostic in this sense; absolute score levels are configuration-specific. We therefore vary both settings in ablations: the frameworks stay below the references across retrieval budgets $k=3,5,10$ (Table~\ref{tab:topk-sweep}) and across answerer models, swapping GPT-4o-mini for GPT-4o or Gemini 2.5 Flash (Table~\ref{tab:answerer-sweep}). Each framework runs under this matched configuration rather than its out-of-the-box default, so that performance differences trace to architectural choices in storage and indexing rather than configuration disparities. Implementation details are in Appendix~\ref{app:adapters}.

\subsection{Evaluation}
\label{sec:eval-protocol}

Session-by-session evaluation reflects how memory systems operate in deployment: each session is ingested once, the system updates its memory state, and queries are answered using only that retained state without the corpus being re-supplied. Sessions are ingested in chronological order; all queries are issued after the final session.

Multiple-choice categories are scored by exact letter match against the correct option. Open-ended categories are scored 0--1 by an LLM judge (GPT-4o-mini) calibrated to avoid verbosity bias: a terse correct answer scores the same as a detailed one. A blinded human check finds substantial agreement with the judge (Appendix~\ref{app:judge-human}). Full judge rubric and per-category scoring rules are in Appendix~\ref{app:harness}.

Results cover all 43 networks across all three tiers: 19 small-tier (264 questions, including 4 networks with member-departure scenarios), 14 medium-tier (346 questions), and 10 large-tier (421 questions, networks ranging 15--30 members). All conditions were evaluated on the same sets of networks at each tier.

\section{Results}
\label{sec:results}

\subsection{Data Quality and Reference Bounds}

Before comparing memory systems, we establish three reference points: whether the answers are recoverable from the conversations, how the matched answerer performs when the memory layer is bypassed, and how raw-turn retrieval performs when no extracted memory state is built. The blind critic, Claude Sonnet 4.5, serves as a validation reference for answer recoverability: it has full conversation access and no planted-challenge metadata, and independently re-answers each question. It reaches a mean score of 0.952. Human reviewers also inspected the released QA pairs in the browser-based audit viewer for evidence grounding, attribution correctness, and answer completeness. Together, these checks show that the gold answers are recoverable from the conversations and supported by the cited evidence. The 0.952 figure is a recoverability gate, not an evaluation oracle: it reports whether a strong reasoning model with full text access can recover each answer, which qualifies the data. The evaluation full-context references are much lower (GPT-4o-mini 0.370 overall; Gemini 2.5 Flash 0.719 against a blind-critic 0.99 on small networks), so the task is hard even with the entire conversation in context.

The two all-network references isolate the answerer and retrieval from memory architecture. The full-context GPT-4o-mini reference reaches 0.370 (memory layer bypassed); the uncompressed raw-turn retrieval reference, also GPT-4o-mini, reaches 0.346 (every turn indexed directly, no extracted memory state).

\subsection{Memory System Performance}
\label{sec:system-perf}

\begin{figure}[t!]
  \centering
  \includegraphics[width=\columnwidth]{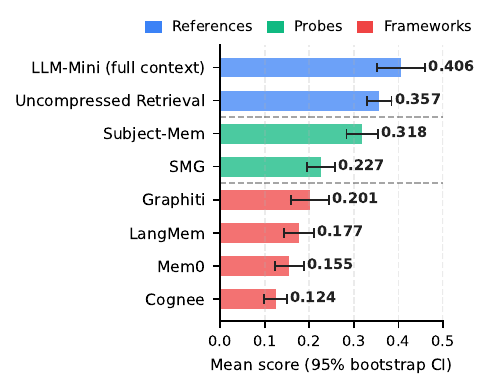}
  \caption{Network-weighted mean score by condition across all 43 evaluated networks, with bootstrap 95\% confidence intervals. All four memory frameworks score below the uncompressed retrieval reference, with overlapping CIs between adjacent frameworks.}
  \label{fig:bar-chart}
\end{figure}

Table~\ref{tab:main-results} and Figure~\ref{fig:bar-chart} show scores across all 43 evaluated networks.

 All four memory frameworks cluster in the 0.12--0.18 question-weighted mean range, well below the uncompressed retrieval reference (0.346). Bootstrap 95\% confidence intervals (10,000 resamples over networks) are shown alongside network-weighted means in Table~\ref{tab:main-results}. Pairwise bootstrap tests show that adjacent frameworks in the ranking are not statistically distinguishable on aggregate. Subject-Mem significantly outperforms the strongest framework, Graphiti ($p < 0.001$). SMG is higher than Graphiti in mean score but not significantly so on aggregate ($p = 0.124$); its main gains appear in specific categories, especially group decisions and norm exceptions. Departed-member retention also separates the systems sharply. Graphiti scores 0.00 on DM, while Mem0 (0.10) and Cognee (0.07) remain near floor. Subject-Mem (0.35), SMG (0.23), and the uncompressed retrieval reference (0.34) retain more departed-member signal, though none solves the category. DM and MA are the smallest categories (10 and 22 questions from 4 and 13 networks) because they need a member departure or a topic the whole group discusses (Appendix~\ref{app:data-construction}), so differences between systems in them carry wider intervals (Table~\ref{tab:category-cis}). Section~\ref{sec:failure-modes} analyzes the architectural mechanisms behind this pattern. Subject-Mem (0.321) nearly matches the uncompressed retrieval reference (0.346) while storing extracted facts rather than every turn. This suggests that much of the frameworks' shortfall may come from their architectural design, in how they extract and index facts, rather than extracted memory itself.

\begin{table*}[t!]
  \centering
  \small
  \resizebox{\textwidth}{!}{%
  \begin{tabular}{lcc cccccccccc}
    \toprule
    \textbf{Condition} & \textbf{Mean$^Q$} & \textbf{Mean$^N$ $\pm$ 95\% CI} & \textbf{SR} & \textbf{GD} & \textbf{MA} & \textbf{AP} & \textbf{TM} & \textbf{NI} & \textbf{RE} & \textbf{TS} & \textbf{DM} \\
    \midrule
    \multicolumn{12}{l}{\textit{Reference conditions (held constant across systems)}} \\
    \textsc{llm-mini} (full context) & 0.370 & $0.406 \pm 0.054$ & 0.34 & 0.66 & 0.12 & 0.75 & 0.31 & 0.59 & 0.23 & 0.28 & 0.38 \\
    \textsc{uncompressed retrieval} & 0.346 & $0.357 \pm 0.027$ & 0.34 & 0.64 & 0.12 & 0.82 & 0.21 & 0.44 & 0.24 & 0.24 & 0.34 \\
    \midrule
    \multicolumn{12}{l}{\textit{Memory frameworks (open-source, current architectural approaches)}} \\
    \textsc{graphiti} & 0.178 & $0.201 \pm 0.043$ & 0.11 & 0.62 & 0.06 & 0.46 & 0.05 & 0.41 & 0.12 & 0.07 & 0.00 \\
    \textsc{langmem} & 0.162 & $0.177 \pm 0.033$ & 0.08 & 0.57 & 0.15 & 0.39 & 0.06 & 0.44 & 0.06 & 0.09 & 0.15 \\
    \textsc{mem0} & 0.143 & $0.155 \pm 0.032$ & 0.11 & 0.58 & 0.05 & 0.31 & 0.05 & 0.24 & 0.06 & 0.07 & 0.10 \\
    \textsc{cognee} & 0.120 & $0.124 \pm 0.025$ & 0.09 & 0.58 & 0.02 & 0.25 & 0.10 & 0.20 & 0.05 & 0.01 & 0.07 \\
    \midrule
    \multicolumn{12}{l}{\textit{Research probes (social-aware indexing ablations)}} \\
    \textsc{subject-mem} & 0.321 & $0.318 \pm 0.035$ & 0.31 & 0.62 & 0.06 & 0.78 & 0.24 & 0.46 & 0.19 & 0.22 & 0.35 \\
    \textsc{smg} & 0.214 & $0.227 \pm 0.031$ & 0.16 & 0.70 & 0.08 & 0.44 & 0.11 & 0.46 & 0.12 & 0.10 & 0.23 \\
    \bottomrule
  \end{tabular}%
  }
  \caption{Mean scores by condition and question category across all 43 evaluated networks. All conditions use GPT-4o-mini as the answerer for matched comparison: \textsc{llm-mini} is the full-context matched-answerer reference, and \textsc{uncompressed retrieval} indexes raw conversation turns without extracted memory state. Both reference rows preserve the original conversation. Mean$^Q$ is question-weighted; Mean$^N$ is network-weighted with bootstrap 95\% confidence intervals from 10,000 resamples over networks. Category codes match Table~\ref{tab:query-types}.}
  \label{tab:main-results}
\end{table*}
\subsection{Performance by Category}

Aggregate scores hide category-level variation. Figure~\ref{fig:heatmap} shows that conditions separate differently by category: attribution probes favor conditions that preserve speaker and subject information, group-decision and norm-exception categories show the clearest gains from the social-graph probe, and theory-of-mind and temporal-shift categories remain low across all evaluated memory frameworks.

Attribution probes show the widest spread. The uncompressed retrieval reference scores 0.82 on AP, and Subject-Mem reaches 0.78; the four memory frameworks span 0.25--0.46. Both high-AP conditions preserve speaker information at ingestion (raw turn text and subject-indexed extraction, respectively); the frameworks compress speaker information at ingestion and lose it. The 0.32-point gain from Subject-Mem over the best framework (0.78 vs. Graphiti 0.46) uses the same answerer and corpus, pointing to ingestion-time indexing as the source of the gain.

SMG's category-level gains isolate the contribution of dissent-aware graph structure. Holding the answerer constant, SMG scores 0.70 on GD against the framework range of 0.57--0.62, and 0.46 on NI, matching the best alternatives (LangMem 0.44, uncompressed 0.44). The gain is recoverable because dissent edges retain information that flat-embedding stores discard at ingestion. Aggregate SMG (0.214) sits below the uncompressed reference (0.346); the result is targeted recovery on graph-structured categories, not overall dominance.

TM and TS stay low across all conditions, including the matched-answerer reference (TM 0.31, TS 0.28). The floor has two compounding causes: the deliberate non-reasoning answerer (Section~\ref{sec:conditions}) and the architectural gaps shared by every evaluated framework. Section~\ref{sec:failure-modes} catalogues the gaps. A matched-oracle comparison against Gemini 2.5 Flash confirms the answerer is not the architectural bottleneck (Appendix~\ref{app:answerer-detail}).

\subsection{Model and Retrieval-Parameter Ablations}

We conducted three ablations to examine how far the results depend on the answering model, the retrieval budget, and extraction quality, all held fixed in the main evaluation. Swapping GPT-4o-mini for GPT-4o or Gemini 2.5 Flash leaves all four frameworks below the references, so a stronger answerer does not close the gap (Table~\ref{tab:answerer-sweep}, Appendix~\ref{app:answerer-detail}). Retrieval budget behaves the same way: as top-$k$ grows from 3 to 10, uncompressed retrieval gains 0.11 while no framework gains more than 0.05, so the frameworks fall further behind raw retrieval instead of catching up (Table~\ref{tab:topk-sweep}, Appendix~\ref{app:topk}). A third ablation swaps Graphiti's extractor from GPT-4o-mini to GPT-4o and finds no overall gain ($-0.036$), with norm-exception unchanged, so extraction quality is not the binding constraint either (Table~\ref{tab:extractor-swap}, Appendix~\ref{app:extractor}).

\section{Failure Mode Analysis}
\label{sec:failure-modes}

The performance gaps in Table~\ref{tab:main-results} trace to specific architectural decisions in each framework. This section pairs each failure mode with the design choice that produces it and the schema primitive that would address it; Table~\ref{tab:failure-modes} summarises all five modes.

Attribution failures have two architectural causes. The first is single-stream conflation: frameworks that ingest all turns into a shared store (Cognee, LangMem, Mem0) flatten ``Sarah said X about Mike'' into a fact about Sarah, or into a group-level summary. Subject-Mem changes only the indexing layer, extracting (subject, fact) pairs and storing facts under the persona they describe; AP rises from the framework range of 0.25--0.46 to 0.78 (Table~\ref{tab:main-results}), pointing to per-persona attribution at ingestion.

The second cause is missing cross-persona representation. Theory-of-mind questions require representing what A revealed about B's preference, not just what B prefers. None of the four frameworks stores this (TM 0.05--0.10). SMG implements a cross-persona edge (\texttt{KNOWS\_ABOUT}) yet scores only 0.11, so representation alone does not close the gap.

Temporal-state failures trace to overwrite-on-update: when a preference changes, the evaluated adapters update the current value without preserving the prior state. Questions of the form ``what drove the change, and what was the prior state?'' cannot be answered from a single current value. Graphiti exposes validity intervals via its bi-temporal model in principle, but the default extraction pipeline does not emit prior-state edges when a new fact arrives. SMG implements an \texttt{EVOLVED\_FROM} edge that links each new fact to the prior state it replaced; SMG still scores 0.10 on TS, and the matched-answerer reference reaches only 0.28, so representation alone does not close the gap. The storage-layer fix is \texttt{EVOLVED\_FROM} plus an extraction step that emits it when a fact contradicts a prior.

Entity merging is the failure mode that most clearly traces to a single framework decision. Graphiti's ingestion pipeline canonicalises personas into shared graph nodes; the large-tier collapse on SR, NI, RE, and TS at 15+ members traces to merging cascades, not to per-network configuration. The clearest signature is departed-member retention: Graphiti scores 0.000 on DM across all 4 evaluated DM networks because subsequent ingestion overwrites the merged-out persona's facts. Subject-Mem and the uncompressed retrieval reference handle DM (0.35 and 0.34 respectively) because neither merges or summarises persona state at ingestion.

Norm-individual conflation surfaces when a group norm has an exception that requires per-persona representation. SMG's dissent-aware schema raises GD to 0.70 against the framework range of 0.57--0.62 and NI to 0.46 against LangMem's 0.44, pointing to the schema layer.

Our research probes target two of the five modes (Subject-Mem on AP; SMG on GD/NI); the other three remain open.

Performance also varies with network size in a U-shape across tiers; the full tier breakdown is in Appendix~\ref{app:tiers}.

\section{Conclusion}

SocialMemBench exposes a consistent gap between what the evaluated memory frameworks achieve and what social group settings require. Across 43 networks, the four evaluated frameworks cluster at 0.12--0.18, well below the uncompressed retrieval reference (0.346) and the matched-answerer full-context reference (0.370). The gaps concentrate on attribution, theory of mind, temporal shift, norm exceptions, and departed-member retention, each a capability an AI assistant in a group chat needs. The benchmark makes social-memory architecture a tractable architectural research problem: each of the five failure patterns is a testable architectural hypothesis, two evidenced in this work by Subject-Mem and SMG, and three that define a research agenda.

Closing the full gap requires per-persona attribution slots, temporally-ordered preference histories, entity-lifecycle preservation, cross-persona knowledge edges, and dissent-aware group-norm schemas. Two are reachable as adapter-layer changes today, one requires extraction-pipeline and storage-schema work, and two require framework-level changes. SocialMemBench provides the measurement to develop these fixes and study social-memory architectures. Closing this gap is a precondition for deploying these systems reliably in social group settings.\textsuperscript{\ref{fn:urls}}

\section*{Limitations}
\label{sec:limitations}

We see several directions for future work. The synthetic generation pipeline trades register variation and naturalism for controlled experimental variables: planted-challenge ground truth at known (speaker, session, turn) coordinates, controlled distribution of rare attribution patterns (theory-of-mind reveals, member-departure dynamics with retained-state probes), and verifiable evidence anchors. Real-data extension is a natural complement, capturing what synthetic generation cannot while requiring different methodological tools to handle the absence of ground truth. Our pipeline models per-persona communication profiles, implicit preferences, relationship histories, and group norms (the structural elements our failure modes target).

The evaluation methodology reflects deliberate tradeoffs. The QA quality check shares a model family with the data generator (Claude Sonnet 4.5 and Claude Code); per-pair human review via the audit viewer (Appendix~\ref{app:viewer}) and downstream evaluation across different model families (GPT-4o-mini in Section~\ref{sec:results}, Gemini 2.5 Flash in Section~\ref{sec:reasoning-comparison}) bound the family-overlap risk, with independent annotation a natural extension. Memory-system evaluations use a single answering model (Section~\ref{sec:conditions}) to isolate memory-layer effects from reasoning capacity; the small-tier oracle comparison (Section~\ref{sec:reasoning-comparison}) shows architectural failure modes are answerer-independent, and extending to stronger answerers would need to control for chain-of-thought masking the architectural failures we measure. Open-ended scoring uses a single GPT-4o-mini judge calibrated against verbosity bias \citep{zheng2023judging}, attribution error, and non-answers; a blinded human check shows substantial agreement (Appendix~\ref{app:judge-human}), and multiple-choice questions (21\% of the dataset) are scored by exact string matching. The benchmark covers English-only text conversation with group sizes 4--30, holding language and modality constant to isolate architectural effects. The architectural failure modes apply at any scale where the underlying framework designs are deployed; multilingual and multimodal (memes, images, voice) extensions are natural follow-ups. We evaluate four open-source memory frameworks (Mem0, LangMem, Graphiti, Cognee) and two research probes, a small sample of possible architectures, so our quantitative results are specific to these systems.

\bibliography{refs}

\appendix

\section{Dataset Construction Details}
\label{app:data-construction}

Generation runs in three stages, each an LLM pass with rule-based verification between stages. Stage 1 builds the network in two passes: a skeleton pass fixes personas, relationship edges, and group norms, and an enrichment pass fills each persona's occupation, skills, implicit preferences, preference-change history, and speaking quirks; the constraints below are checked before the network proceeds. Stage 2 plans every session first (date, topic, active members, and two to four planted challenges), derives a voice fingerprint per persona, writes each session as a chat thread, and then anchors each planted challenge to the specific turn and verbatim span that expresses it. Stage 3 generates and verifies the QA pairs (Section~\ref{sec:data-gen}). No real-world data enters at any stage.

\subsection{Network Generation: Schema and Constraints}

Each network is stored as a JSON file (\texttt{data/raw/\{network\_id\}.json}) with four top-level structures. The \texttt{personas} field is a list of $n$ persona objects, each carrying display name, age, occupation, background, Big Five personality scores on a three-point scale, a seven-dimension communication profile (formality, verbosity, emoji use, humor, directness, typo rate, and warmth), categorized preferences, a preference history recording old value, new value, trigger, and session index per event, and two to three distinctive speaking quirks. The \texttt{edges} field records pairwise relationship edges with relationship type, closeness, sentiment, a history summary, and shared events. The \texttt{group\_norms} field lists group-level behavioral norms, each with a boolean \texttt{truly\_universal} flag and a dissenters list that serves as the Q6 ground truth anchor. The \texttt{metadata} field records sessions planned, turns per session, and departure events.

The following structural constraints are enforced at Stage 1 and verified before any network proceeds to Stage 2. Closeness ratings are not uniformly high: at least one relationship edge must carry mixed or tense sentiment. At least one group norm must be non-universal with a non-empty dissenters list. Every persona must have at least one preference history event. No two personas may share the same formality-emoji-humor triple, and every persona pair must differ on at least two of verbosity, emoji use, and formality. Edge coverage is enforced by group size: five-person networks require at least six edges; nine-person networks require at least twelve.

\subsection{Conversation Generation: Planted Challenge Types}

Seven planted challenge types are distributed across sessions during Stage 2 generation; an eighth, member departure, is encoded at the network level rather than as a planted conversational signal (Table~\ref{tab:challenge-types}). Each planted challenge records: \texttt{challenge\_type}, \texttt{subject\_persona\_id}, \texttt{speaker\_persona\_id}, \texttt{fact\_attribute}, \texttt{fact\_value}, \texttt{evidence\_span} (verbatim substring of the message, $\leq$200 chars), \texttt{implicitness\_level} (1--3), \texttt{session\_index}, and \texttt{turn\_index}.

\begin{table*}[t!]
  \centering
  \small
  \begin{tabular}{p{3.8cm} p{4.5cm} p{3.5cm}}
    \toprule
    \textbf{Challenge type} & \textbf{Definition} & \textbf{Seeds QA type(s)} \\
    \midrule
    \texttt{implicit\_preference} & Subject's preference revealed through behavior or complaint, not stated directly & SR, MA \\
    \texttt{consensus\_trap} & Group makes a decision; one dissenter goes quiet or deflects & GD, NI \\
    \texttt{false\_attribution\_seed} & Two personas make similar-sounding statements on related topics in distinct voices & AP \\
    \texttt{theory\_of\_mind} & Speaker A reveals what they know about Speaker B's preference via an observable action & TM \\
    \texttt{temporal\_shift} & Persona's preference contradicts an earlier stated preference; change has a trigger & TS \\
    \texttt{relational\_disclosure} & History or nature of a relationship between two personas emerges in passing & RE \\
    \texttt{relational\_graph} & Group structure revealed: how members met, sub-clique dynamics, closeness signals & RE \\
    \texttt{member\_departure} & Member leaves the network mid-corpus; pre-departure facts become Q9 ground truth & DM \\
    \bottomrule
  \end{tabular}
  \caption{Planted challenge types and their role in the QA ground truth. Codes match Table~\ref{tab:query-types}. The final row (member departure) is structural, encoded at the network level rather than as a planted conversational signal.}
  \label{tab:challenge-types}
\end{table*}
The per-category counts are uneven, as one would expect when these phenomena do not occur at equal rates in group conversation. Some challenge types are planted more often than others, departed-member questions come only from networks with a departure, and multi-contact aggregation requires a whole-group domain, so DM (10) and MA (22) are the smallest categories.

\begin{table}[t]
  \centering \small
  \begin{tabular}{lr}
    \toprule
    \textbf{Category} & \textbf{$n$} \\
    \midrule
    SR (single-contact recall)     & 249 \\
    GD (group decision)            & 81 \\
    MA (multi-contact aggregation) & 22 \\
    AP (attribution probe)         & 80 \\
    TM (theory of mind)            & 92 \\
    NI (norm vs.\ individual)      & 54 \\
    RE (relational edge)           & 181 \\
    TS (temporal shift)            & 262 \\
    DM (departed member)           & 10 \\
    \midrule
    Total                          & 1{,}031 \\
    \bottomrule
  \end{tabular}
  \caption{QA pairs per category across the 43 released networks (1{,}031 total).}
  \label{tab:category-counts}
\end{table}
Implicitness levels control difficulty. Level 1 preferences are stated directly and are used sparingly, mostly as foil setups for attribution probes. Level 2 (the default) requires inferring the preference from a complaint, reaction, or passing comment. Level 3 (the hardest) requires reading silence, deflection, or subject change as a preference signal.

\subsection{Conversation Quality Acceptance Criteria}

The Stage 2 QC blind-critic pass scores five conversation quality dimensions on a 0--2 scale. A network is accepted only when all five dimensions score $\geq 1$ and \texttt{em\_dash\_free} scores exactly 2.

\begin{table*}[t!]
  \centering
  \small
  \begin{tabular}{p{3.0cm} p{6.8cm} p{1.8cm}}
    \toprule
    \textbf{Dimension} & \textbf{Pass criteria} & \textbf{Score 0 trigger} \\
    \midrule
    Topic diversity & Sessions cover topics consistent with the group type's real behavior (recreational: activity-anchored; close-friends: distinct per-session thread; family: rotating domains) & Any session is a near-duplicate of another \\
    Voice consistency & Per-persona voice fingerprints realized across 5 sub-dimensions: emoji use, typo rate, verbosity, formality, speaking quirks & Any persona's emoji or verbosity contradicts their profile \\
    Transition naturalness & Topic shifts triggered by a character raising something; no cold jumps or announced pivots & Any unexplained topic switch \\
    Em-dash free & Zero \texttt{---} in any \texttt{message} field (generation artifact absent in real chat) & Any em dash found \\
    Planted turn integrity & All evidence spans are verbatim substrings of the corresponding turn message & Any span not found verbatim \\
    \bottomrule
  \end{tabular}
  \caption{Conversation quality dimensions assessed in the Stage 2 QC pass.}
  \label{tab:qc-dimensions}
\end{table*}
\subsection{QA Generation: Inference Depth Rubric}

The inference depth rubric is applied as a pre-generation gate. Any pair that fails is either reframed or skipped.

\begin{table*}[t!]
  \centering
  \small
  \begin{tabular}{p{0.8cm} p{2.0cm} p{6.7cm}}
    \toprule
    \textbf{Depth} & \textbf{Label} & \textbf{Definition} \\
    \midrule
    0 & Reject & Answerable by keyword or name search: one word in the transcript gives the answer \\
    1 & Easy & Single-turn contextual reading required; answer is stated in one place \\
    2 & Medium & Two-turn synthesis or one inference step; connecting two pieces of evidence \\
    3 & Hard & Cross-session inference or attribution ambiguity; tracking signals across $\geq$3 turns or distinguishing two plausible speakers by meaning \\
    \bottomrule
  \end{tabular}
  \caption{Inference depth rubric used to assign difficulty labels and filter trivially answerable pairs.}
  \label{tab:inference-rubric}
\end{table*}
\subsection{Query-Type Structural Flags}

The Stage 3 QC pass applies rule-based structural flags per query type; flagged pairs are rewritten or removed. General flags reject pairs that are trivially answerable by keyword search or that embed the answer (a preference keyword or speaker name) in the question text. Type-specific flags enforce structural requirements: Q4 pairs must include a contamination foil with at least a two-sentence explanation naming a specific vocabulary or framing similarity between target and foil speaker; Q5 pairs must ask both what was revealed and who revealed it through observable chat behavior (not a claimed private conversation); Q8 pairs must have at least two temporal anchors, preferably in non-consecutive sessions, and must ask for the trigger as well as the change.

\subsection{Dataset File Format}
{\sloppy

Conversations (\texttt{data/conversations/}\allowbreak\texttt{\{id\}.json}): JSON with \texttt{sessions} list; each session has \texttt{session\_id}, \texttt{date}, \texttt{date\_label}, \texttt{topic}, and \texttt{turns}. Each turn has \texttt{turn\_id}, \texttt{speaker\_display\_name}, \texttt{timestamp}, \texttt{message}, and \texttt{planted\_challenges} (list of challenge objects). A flat \texttt{planted\_challenge\_index} at the corpus root allows fast lookup without traversing all sessions.

QA pairs (\texttt{data/qa/\{id\}.jsonl}): one JSON object per line. Key fields: \texttt{qa\_id}, \texttt{query\_type}, \texttt{difficulty}, \texttt{question}, \texttt{answer}, \texttt{answer\_format} (long\_form / multiple\_choice / short\_answer), \texttt{options}, \texttt{correct\_option}, \texttt{evidence\_anchors} (list of session index + turn ID + verbatim excerpt), \texttt{contamination\_foil} (Q4 only), \texttt{temporal\_anchors} (Q8 only), and QC fields: \texttt{qc\_phase1\_score}, \texttt{qc\_phase2\_grounded}, \texttt{qc\_phase3\_flagged}.

Manifest (\texttt{data/manifest.json}): incremental tracking of all networks; records status, session count, QA count, consumed challenge IDs, eval summary, and verify status per network.
\par}

\subsection{Answerability-Threshold Sensitivity}

The 0.70 blind-critic gate is a floor the released data clears by a wide margin, so admission is insensitive to the exact value. The mean blind-critic score is 0.952 and every network mean is $\geq 0.86$, so all 43 networks also clear the network-level gate. Table~\ref{tab:threshold-sensitivity} sweeps the per-pair gate.

\begin{table}[t]
  \centering \small
  {
  \begin{tabular}{cc}
    \toprule
    Per-pair gate & Pairs retained \\
    \midrule
    0.70 & 100.0\% \\
    0.75 & 99.8\% \\
    0.80 & 99.0\% \\
    0.85 & 97.6\% \\
    0.90 & 92.5\% \\
    \bottomrule
  \end{tabular}}
  \caption{Sensitivity to the per-pair answerability gate: fraction of the 1{,}031 QA pairs retained as the threshold is raised. The 0.70 gate is a floor the data clears by a wide margin (mean blind-critic score 0.952).}
  \label{tab:threshold-sensitivity}
\end{table}

\subsection{Worked Example: From Network to QA Pair}
\label{app:worked-example}

{
This example follows one planted fact through the three stages of Section~\ref{sec:data-gen}, using the released records for network \texttt{grp\_d3e4f5a6}, a 22-member close-friends network in the large tier. Excerpts are abridged with [\ldots].

\paragraph{Stage 1 instruction.} The network-generation prompt requires every persona to carry a preference change tied to a planned session:
\begin{verbatim}
preference_history: >=1 event per persona
  [...] must include a session
reference within the planned session range
[ ] Every persona has >=1 preference_history
  event (Q8 ground truth)
\end{verbatim}

\paragraph{Stage 1 output.} Priya's persona record and two of her 3 relationship edges in the 22-persona network:
\begin{verbatim}
{"persona_id": "p_d3e40003", "display_name":
  "Priya",
 "occupation": "Hospital pharmacist",
 "communication_profile": {"formality":
   "Semi-formal",
   "verbosity": "Terse", "directness":
     "Direct", ...},
 "preference_history": [{"attribute":
   "exercise_pattern",
   "old_value": "ran half-marathons and posted
     training plans to the chat",
   "new_value": "switched to volleyball and
     almost never mentions running",
   "changed_at_session": 3,
   "trigger": "Stress fracture during her
     residency forced a switch;
               she has since fallen in love
                 with team sport ..."}]}

{"persona_a_id": "p_d3e40001" (Maya),
  "persona_b_id": "p_d3e40003" (Priya),
 "relationship_type":
   "research_methods_classmates", "closeness":
   "High"}
{"persona_a_id": "p_d3e40003" (Priya),
  "persona_b_id": "p_d3e40004" (Tomas),
 "relationship_type":
   "volleyball_league_partners", "closeness":
   "High",
 "history_summary": "Priya and Tomas run a
   volleyball league together; ..."}
\end{verbatim}

\paragraph{Stage 2 instruction.} The conversation-generation prompt defines the challenge type that carries this fact:
\begin{verbatim}
temporal_shift: A persona expresses a
  preference that CONTRADICTS their
earlier preference (from a prior session). The
  shift should feel earned
[...] This is ground truth for Q8.
\end{verbatim}

\paragraph{Stage 2 output.} The planted-challenge record and the session-3 turns it points to (topic: ``Health, exercise routines, and a long thread about training''):
\begin{verbatim}
{"challenge_id": "ch_3fb0515d",
  "challenge_type": "temporal_shift",
 "subject_persona_id": "p_d3e40003",
   "speaker_persona_id": "p_d3e40003",
 "fact_value": "Priya shifted from running
   half-marathons to volleyball
                after a stress fracture",
 "evidence_span": "stress fracture during
   residency. team sport now.
                   easier on the body.",
 "implicitness_level": 3, "session_index": 3}

[Priya]: joining the volleyball league
         finals this saturday if anyone
         wants to come watch.
[...]
[Devon]: wait priya did you used to do half
  marathons? i remember the
         training plan posts
[Priya]: i did. moved on.
[Aanya]: lol there is a story behind 'moved
  on' but priya is not gonna
         give it to you in this chat devon
[Priya]: stress fracture during residency.
  team sport now. easier on
         the body.
\end{verbatim}

\paragraph{Stage 3 output.} The QA pair seeded by this challenge (\texttt{Q8\_75d136f5}, temporal shift, medium difficulty, open-ended), which appears in Table~\ref{tab:query-types}:
\begin{verbatim}
Q: What in the conversation signals that
  Priya's relationship to running
   and team sport has shifted, and what
     appeared to drive it?
A: Early state: Priya used to post training
  plans and ran half-marathons
   [...] Shift signal: Priya replies with
     two-word answers, 'i did.' and
   'moved on.' [...] Trigger: a stress
     fracture during residency forced
   the change [...] Evolved state: She is now
     firmly volleyball-focused
   [...] The replacement is named, not just
     the loss.
Evidence: session 3, turns
  grp_d3e4f5a6_s03_t006 and _t011
\end{verbatim}
The pair passed every gate in Section~\ref{sec:data-gen}: no keyword shortcut answers it (inference depth 2), its evidence span matches the corpus verbatim, the blind critic scored it 0.93, and it was approved in human review in the audit viewer.
}

\section{Memory System Adapter Details}
\label{app:adapters}

All memory system evaluations follow a session-sequential ingestion protocol: sessions are processed in chronological order; the memory system is not shown future sessions during ingestion. Queries are issued after the final session has been ingested.

\subsection{Adapter Interface}

All adapters implement a common Python abstract base class (\texttt{eval/adapters/base.py}) with two async methods: \texttt{add()} ingests all sessions into the memory system, and \texttt{search()} retrieves relevant memories for a given question. Each system has its own adapter because the libraries expose different APIs; every adapter drives that system's native ingestion and retrieval without reimplementing its internals. The retrieval budget $k$ and the answerer model are varied only in the ablations (Appendices~\ref{app:answerer-detail} and~\ref{app:topk}).

The adapter's \texttt{search()} output is passed verbatim as \texttt{\{context\}} in the answerer prompt. This means retrieval quality and formatting directly affect the answering model's performance.

\subsection{Adapter Configurations}

Table~\ref{tab:adapters} summarizes each adapter's implementation, Python environment, and key design choices.

\begin{table*}[t!]
  \centering
  \small
  \begin{tabular}{p{2.2cm} p{2.8cm} p{1.4cm} p{2.5cm} p{3.9cm}}
    \toprule
    \textbf{Condition} & \textbf{Implementation} & \textbf{Env} & \textbf{Extraction} & \textbf{Retrieval} \\
    \midrule
    \textsc{uncompressed retrieval} & Chroma vector store, raw turns verbatim & smb310 & None (raw text) & Cosine similarity; \texttt{text-embedding-3-small}; top-10 \\
    \textsc{subject-mem} & Chroma; indexes facts by subject, not speaker & smb310 & GPT-4o-mini & Cosine similarity; \texttt{text-embedding-3-small}; top-10 \\
    \textsc{smg} & Social Memory Graph (JSON-backed graph) & smb310 & GPT-4o-mini & Query-type-routed graph traversal; top-10 \\
    \textsc{graphiti} & Graphiti-core 0.28.2 + Kuzu 0.11.3 & smb310 (Py 3.10) & GPT-4o-mini & FTS + graph; top-10 \\
    \textsc{langmem} & LangChain InMemoryStore + reflection & smb311 (Py 3.11) & GPT-4o-mini & Semantic search; top-10 \\
    \textsc{mem0} & Mem0 local + Chroma & smb310 & GPT-4o-mini & Semantic; top-10 \\
    \textsc{cognee} & Cognee batch doc-to-graph & smb310 (Py 3.10) & GPT-4o-mini & Hybrid graph+vector \\
    \midrule
    \textsc{llm-mini} & Full context (no memory system) & smb310 & n/a & Full corpus \\
    \textsc{llm-gemini} & Full context (no memory system) & smb310 & n/a & Full corpus \\
    \bottomrule
  \end{tabular}
  \caption{Memory system adapter configurations. All adapters use GPT-4o-mini as the answering model; extraction model column refers to memory ingestion only.}
  \label{tab:adapters}
\end{table*}
\subsection{Adapter Implementation Notes}

\textsc{Subject-Mem} uses the same Chroma vector store as the uncompressed retrieval baseline but changes the unit of storage. At ingestion, GPT-4o-mini parses each session turn and extracts (subject, fact) pairs, where the subject is the person the fact is \emph{about}, not the person who said it. Each pair is stored as ``[About: \{subject\}] \{fact\}'' with speaker metadata. This transforms the indexing from speaker-anchored to subject-anchored without changing the retrieval mechanism, isolating the contribution of ingestion-time attribution.

\textit{Uncompressed retrieval baseline.} It stores each conversation turn as a single Chroma document with metadata fields \texttt{speaker}, \texttt{session\_index}, and \texttt{timestamp}. Retrieval is purely cosine similarity against the question string; no LLM call is made during ingestion. Because raw turn text contains speaker labels verbatim, this baseline scores 0.82 on attribution probes by string matching rather than social reasoning. It is reported as a reference, not as a memory system.

\textit{Social Memory Graph (SMG).} It maintains a structured JSON graph with node types: \texttt{PERSONA} (per-member), \texttt{PREFERENCE} (attribute-value facts), \texttt{GROUP\_NORM} (group-level facts), and \texttt{SESSION}. Edge types include: \texttt{HAS\_PREFERENCE}, \texttt{DISSENTS\_FROM} (norm exceptions), \texttt{KNOWS\_ABOUT} (cross-persona knowledge), \texttt{EVOLVED\_FROM} (superseded preferences), and \texttt{PARTICIPATED\_IN}. Query-type routing in \texttt{eval/adapters/smg/queries.py} directs Q2 and Q6 queries to \texttt{DISSENTS\_FROM} traversal and Q5 queries to \texttt{KNOWS\_ABOUT} traversal.

\textit{Graphiti} uses the Graphiti-core library with an embedded Kuzu graph database (no Docker required). During ingestion, each session's turns are submitted as episodes; Graphiti extracts entity-relationship triples using GPT-4o-mini. Two implementation-level bugs were fixed before evaluation: (i) \texttt{KuzuDriver.\_database} was never set, causing an \texttt{AttributeError} on \texttt{add\_episode()}. Fixed by setting \texttt{driver.\_database = network\_id} after \texttt{KuzuDriver} creation; (ii) Kuzu FTS indexes were not created by the default \texttt{build\_indices\_and\_constraints()} in version 0.28.2. Fixed by adding an explicit \texttt{CREATE\_FTS\_INDEX} call for all relevant tables.

\textit{LangMem} uses LangChain's \texttt{InMemoryStore} with a reflection step that consolidates memories after each session. An embedding error in the store's \texttt{search()} method silently returned empty results in some configurations. Fixed by adding explicit exception logging and a fallback that returns all memories as flat context.

\textit{Mem0 (local)} uses the Mem0 Python SDK with local Chroma as the vector backend. An early bug caused per-turn ingestion to strip speaker attribution (storing ``User likes X'' without the speaker name). Fixed by batch-ingesting whole sessions using the Mem0 messages API, which preserves the \texttt{role}/\texttt{content} structure.

\section{Evaluation Pipeline Details}
\label{app:harness}

\subsection{CLI and Pipeline}

The evaluation pipeline is invoked via:

\begin{verbatim}
python -m eval.cli --network {network_id} \
    --condition {condition}
\end{verbatim}

The pipeline executes four stages: (1) load the conversation and QA files; (2) ingest sessions into the memory system via \texttt{adapter.add()}, or assemble full-corpus context for full-context conditions; (3) for each QA pair, call \texttt{adapter.search()} and prompt GPT-4o-mini with the retrieved context and question; (4) score each answer against the ground truth.

Results are written to \path{data/results/{condition}/{network_id}.jsonl} (per-pair) and \path{data/results/{condition}/{network_id}_summary.json} (summary).

\subsection{Answerer Prompts}

Separate prompt templates are used for memory-system conditions (context = retrieved memories) and full-context conditions (context = complete corpus). Both templates include query-type-specific answering guidance. For example, the Q8 guidance instructs the model to state the old preference, the new preference, and the trigger. All three are required for full credit.

The memory-system prompt includes explicit instructions for reading SMG-formatted context: ``Facts labeled `session N' are chronological. `EVOLVED\_FROM' / `Earlier belief (session N)' means the newer fact supersedes it.'' This ensures the prompt design does not artificially disadvantage structured memory outputs.

\subsection{LLM Judge Rubric}

Open-ended answers are scored by GPT-4o-mini acting as a judge. The judge receives the question, the gold answer, and the generated answer, and returns a JSON object \texttt{\{"score": 0.0--1.0, "rationale": "..."\}}. We use a graded rubric rather than a categorical label because several open-ended categories are compositional: Q3 credits the fraction of members recalled and Q8 credits the old state, new state, and trigger separately, so a binary label would score a partially correct answer (say, four of five members) the same as a fully wrong one and discard the gradient that distinguishes systems. The judge maps answers to the anchored bands and rules below, not free-form numbers.

Three critical rules are applied before the general rubric:
\begin{enumerate}
  \item \textbf{Anti-verbosity bias}: a concise correct answer scores the same as a verbose one. Evidence citation is not required for Q1/Q3. This rule was added after observing that early judge versions systematically downscored terse-but-correct responses.
  \item \textbf{Attribution error}: attributing a fact to the wrong person always scores $\leq 0.3$, regardless of how much other content is correct.
  \item \textbf{Non-answer penalty}: ``NOT ENOUGH INFORMATION'' or ``[No memories retrieved]'' always scores 0.0.
\end{enumerate}

Query-type-specific scoring rules are applied on top. Q3 scores are the fraction of group members correctly recalled. Q5 requires both the preference and who revealed it through observable chat behavior for a score of 1.0; either part alone scores 0.5. Q8 awards one third of a point each for the old state, the new state, and the trigger; both states without a trigger scores 0.7.

Multiple-choice questions (Q2, Q4, Q6) use exact letter matching (binary 1.0/0.0) with a robust parser that handles formats such as ``A'', ``A.'', ``Answer: A'', and ``The answer is A''.

\subsection{Human Agreement with the Judge}
\label{app:judge-human}

Following the meta-evaluation of \citet{longmemeval2025}, one annotator labelled a seeded sample of open-ended answers as correct or incorrect, seeing the question, the gold answer with its evidence, and the system answer, but not the system identity or the judge's score; the judge counts as correct at 0.7, the rubric's boundary for a right answer with at most a minor part missing. On 50 answers, agreement was 84\% (Cohen's $\kappa = 0.66$, 95\% bootstrap CI 0.43--0.86), substantial agreement \citep{landis1977measurement}.

\subsection{Reproducibility Notes}
\label{app:repro}

\begin{enumerate}
  \item All 43 networks and 1,031 QA pairs are generated by Claude Sonnet 4.5 via Claude Code following the deterministic pipeline in Section~\ref{sec:data-gen}; no API key is required. Each network has an integer seed recorded in \texttt{manifest.json}.
  \item Model versions: answering model GPT-4o-mini (\texttt{gpt-4o-mini-2024-07-18}); full-context baselines GPT-4o-mini and Gemini 2.5 Flash (\texttt{gemini-2.5-flash}); judge GPT-4o-mini; SMG and Graphiti extraction GPT-4o-mini. All temperatures are 0.0.
  \item Pipeline concurrency: 10 concurrent judge calls, 5 concurrent answer generation calls. Retry policy: exponential backoff, up to 5 retries, maximum delay 60s.
  \item Environment: macOS Darwin 25.3.0; Python 3.10 (smb310 env, all conditions except LangMem), Python 3.11 (smb311 env, LangMem).
  \item The dataset is at \url{https://huggingface.co/datasets/qunlexie/SocialMemBench} (CC BY 4.0); the evaluation pipeline, generation skills, and browser-based data audit viewer are at \url{https://github.com/Qunlexie/SocialMemBench} (MIT). The repository includes a \texttt{requirements.txt} and per-adapter environment specifications. All memory system adapter patches applied before evaluation are included in the repository and described in Appendix~\ref{app:adapters}.
  \item Exact software versions for evaluated systems: Graphiti-core 0.28.2 with Kuzu 0.11.3; Mem0 (\texttt{mem0ai}) 1.0.11 with local Chroma 1.5.7; LangMem 0.0.30 with LangGraph 1.1.6 \texttt{InMemoryStore}; Cognee 0.5.8 (batch doc-to-graph); OpenAI Python SDK 2.31.0. Evaluation was conducted April 2026.
\end{enumerate}

\section{Answerer Model Comparison}
\label{app:answerer-detail}
\label{sec:reasoning-comparison}

GPT-4o-mini is the answerer for all conditions in Table~\ref{tab:main-results}. Most deployed memory systems run on smaller models; the matched-answerer choice mirrors that and isolates memory-layer effects from reasoning capacity. Table~\ref{tab:reasoning-comparison} reports a matched comparison against Gemini 2.5 Flash on the same 10 small-tier networks (77 questions), full-context oracle for both with the same GPT-4o-mini judge. Gemini answers; it does not index or extract.

\begin{table*}[t!]
  \centering
  \small
  \begin{tabular}{lccccccccc}
    \toprule
    \textbf{Model} & \textbf{Mean} & \textbf{SR} & \textbf{GD} & \textbf{MA} & \textbf{AP} & \textbf{TM} & \textbf{NI} & \textbf{RE} & \textbf{TS} \\
                   & ($n$=77) & (15) & (5) & (5) & (8) & (10) & (8) & (10) & (16) \\
    \midrule
    GPT-4o-mini (full ctx) & 0.644 & 0.67 & 0.60 & 0.34 & 0.50 & 0.79 & 1.00 & 0.65 & 0.53 \\
    Gemini 2.5 Flash (full ctx) & 0.719 & 0.76 & 0.60 & 0.64 & 0.62 & 0.83 & 1.00 & 0.68 & 0.61 \\
    \bottomrule
  \end{tabular}
  \caption{Answerer-model comparison: per-category mean scores on 10 small-tier networks (77 questions), full-context oracle for both models, same GPT-4o-mini judge. DM (Q9) is omitted because the 10-network subset does not contain member-departure scenarios; DM scores are reported in Table~\ref{tab:main-results}.}
  \label{tab:reasoning-comparison}
\end{table*}
The aggregate gap is small: 0.075 in Gemini's favour. Per-category gaps are smaller still, under 0.13 on every category except Q3 (MA: +0.30 on $n=5$, noisy). TM and TS, the categories where memory frameworks score lowest, show gaps of +0.04 and +0.08. GPT-4o-mini reaches TM 0.79 on small-tier; the TM 0.31 in Table~\ref{tab:main-results} reflects the medium- and large-tier rows. The answerer model is not the architectural bottleneck.

Table~\ref{tab:answerer-difficulty} gives the difficulty-stratified breakdown on the same 10 small-tier networks (77 questions), full-context oracle for both models, with the same GPT-4o-mini judge.

\begin{table*}[t!]
  \centering
  \small
  \begin{tabular}{lcccc}
    \toprule
    \textbf{Model} & \textbf{Mean} & \textbf{Easy ($n=6$)} & \textbf{Medium ($n=57$)} & \textbf{Hard ($n=14$)} \\
    \midrule
    GPT-4o-mini (full ctx) & 0.644 & 0.95 & 0.66 & 0.44 \\
    Gemini 2.5 Flash (full ctx) & 0.719 & 0.97 & 0.73 & 0.58 \\
    \bottomrule
  \end{tabular}
  \caption{Difficulty-stratified breakdown of the answerer-model comparison. Difficulty levels follow the inference-depth rubric (Section~\ref{sec:data-gen}): easy = single-turn lookup, medium = 2-turn synthesis or one inference step, hard = cross-session inference or attribution ambiguity.}
  \label{tab:answerer-difficulty}
\end{table*}
Table~\ref{tab:reasoning-comparison} swaps the answerer only for the full-context oracle. Table~\ref{tab:answerer-sweep} extends the swap to every condition, keeping the stored memory, retrieval, and judge fixed and varying only the answering model.

\begin{table*}[t!]
  \centering \small
  {
  \begin{tabular}{lccc}
    \toprule
    \textbf{Condition} & \textbf{GPT-4o-mini} & \textbf{GPT-4o} & \textbf{Gemini 2.5 Flash} \\
    \midrule
    \textsc{llm-mini} (full ctx)   & $0.402 \pm 0.055$ & $0.465 \pm 0.067$ & $0.384 \pm 0.057$ \\
    \textsc{uncompressed retrieval} & $0.362 \pm 0.047$ & $0.395 \pm 0.055$ & $0.347 \pm 0.058$ \\
    \textsc{subject-mem}           & $0.290 \pm 0.054$ & $0.345 \pm 0.058$ & $0.364 \pm 0.081$ \\
    \textsc{smg}                   & $0.242 \pm 0.064$ & $0.268 \pm 0.055$ & $0.267 \pm 0.068$ \\
    \textsc{graphiti}              & $0.230 \pm 0.066$ & $0.217 \pm 0.057$ & $0.221 \pm 0.064$ \\
    \textsc{langmem}               & $0.222 \pm 0.057$ & $0.262 \pm 0.064$ & $0.234 \pm 0.074$ \\
    \textsc{mem0}                  & $0.171 \pm 0.075$ & $0.172 \pm 0.051$ & $0.137 \pm 0.054$ \\
    \textsc{cognee}                & $0.132 \pm 0.035$ & $0.066 \pm 0.038$ & $0.083 \pm 0.040$ \\
    \bottomrule
  \end{tabular}
  }
  \caption{Answerer sweep over the fixed memory state: network-weighted mean $\pm$ 95\% CI (bootstrap over the 15 networks) as the answering model is swapped (same stored memory, retrieval, and judge), ablation subset (174 questions), $k{=}10$. Question-weighted means agree within 0.04 (released CSV). A stronger answerer lifts the conditions that retain the fact but leaves every framework below the references.}
  \label{tab:answerer-sweep}
\end{table*}
Extending the swap to the memory systems isolates where answering capacity matters. Over the same stored memory, no framework reaches the uncompressed-retrieval reference under any answerer. A stronger answerer cannot recover a fact that ingestion did not store, so the binding constraint is the memory layer, not the answering model.

\section{Retrieval-Budget Sensitivity}
\label{app:topk}

To test whether the comparison depends on the retrieval budget, we vary top-$k$ with the answerer (GPT-4o-mini) and judge fixed, on the ablation subset (15 networks, 174 questions). top-$k$ is applied uniformly and acts only at retrieval, so each memory state is built once and reused across $k$.

\begin{table*}[t!]
  \centering \small
  {
  \begin{tabular}{lccc}
    \toprule
    \textbf{Condition} & \textbf{$k{=}3$} & \textbf{$k{=}5$} & \textbf{$k{=}10$} \\
    \midrule
    \textsc{uncompressed retrieval} & $0.254 \pm 0.066$ & $0.304 \pm 0.063$ & $0.362 \pm 0.047$ \\
    \textsc{subject-mem}            & $0.267 \pm 0.053$ & $0.289 \pm 0.067$ & $0.290 \pm 0.054$ \\
    \textsc{smg}                    & $0.220 \pm 0.045$ & $0.242 \pm 0.053$ & $0.242 \pm 0.064$ \\
    \textsc{graphiti}               & $0.200 \pm 0.063$ & $0.201 \pm 0.066$ & $0.230 \pm 0.066$ \\
    \textsc{langmem}                & $0.218 \pm 0.046$ & $0.222 \pm 0.058$ & $0.222 \pm 0.057$ \\
    \textsc{mem0}                   & $0.143 \pm 0.048$ & $0.169 \pm 0.059$ & $0.171 \pm 0.075$ \\
    \textsc{cognee}                 & $0.087 \pm 0.033$ & $0.091 \pm 0.039$ & $0.132 \pm 0.035$ \\
    \bottomrule
  \end{tabular}
  }
  \caption{Retrieval-budget sweep: network-weighted mean $\pm$ 95\% CI (bootstrap over the 15 networks) as top-$k$ varies (answerer GPT-4o-mini and judge fixed), ablation subset (174 questions). Question-weighted means agree within 0.04. No framework reaches the uncompressed-retrieval reference at any $k$; the full-context oracle is $\sim$0.40 on this subset.}
  \label{tab:topk-sweep}
\end{table*}
Larger budgets help modestly (uncompressed retrieval rises from 0.254 to 0.362), but no framework reaches the uncompressed-retrieval reference at any $k$, and all stay well below the full-context oracle. Because top-$k$ acts after ingestion, more budget can only surface facts that were stored, not recover ones that were mis-indexed or overwritten at ingestion, so the comparison does not hinge on the $k{=}10$ choice.

\section{Extractor-Model Ablation}
\label{app:extractor}

The answerer sweep (Appendix~\ref{app:answerer-detail}) shows the answering model is not the bottleneck. This ablation asks the complementary question on the write side: does a stronger extraction model close the gap? We replace Graphiti's default GPT-4o-mini extractor with GPT-4o, holding the answerer (GPT-4o-mini), retrieval, and judge fixed, on the ablation subset (15 networks, 174 questions). We use Graphiti because it is the strongest framework (Table~\ref{tab:main-results}), the most favourable case for an upgrade, and it exposes its extractor as a configuration setting, so the swap changes extraction alone.

\begin{table}[t]
  \centering \small
  \setlength{\tabcolsep}{4pt}
  {
  \begin{tabular}{lrrrr}
    \toprule
    \textbf{Category} & \textbf{$n$} & \textbf{4o-mini} & \textbf{4o} & \textbf{$\Delta$} \\
    \midrule
    SR (recall)            & 33 & 0.179 & 0.070 & $-0.109$ \\
    GD (group decision)    & 13 & 0.462 & 0.462 & $0.000$ \\
    MA (aggregation)       &  5 & 0.260 & 0.200 & $-0.060$ \\
    AP (attribution)       & 19 & 0.526 & 0.632 & $+0.105$ \\
    TM (theory of mind)    & 20 & 0.090 & 0.000 & $-0.090$ \\
    NI (norm-exception)    & 12 & 0.500 & 0.500 & $0.000$ \\
    RE (relational edge)   & 29 & 0.172 & 0.131 & $-0.041$ \\
    TS (temporal)          & 43 & 0.114 & 0.084 & $-0.030$ \\
    \midrule
    All                    & 174 & 0.235 & 0.199 & $-0.036$ \\
    \bottomrule
  \end{tabular}
  }
  \caption{Extractor-model ablation on Graphiti: per-category mean score with the default GPT-4o-mini extractor versus a GPT-4o extractor (answerer, retrieval, and judge fixed), ablation subset (15 networks, 174 questions). DM does not occur in the subset. A stronger extractor does not improve overall performance; the only gain is attribution (AP). The norm-exception (NI) and temporal-shift (TS) categories, where the failure-mode analysis locates the loss, do not improve.}
  \label{tab:extractor-swap}
\end{table}

The upgrade does not help Graphiti overall ($-0.036$ question-weighted; $-0.063$ on the per-network mean, 0.207 versus 0.270). The only category that improves is attribution (AP, $+0.105$), where subject attribution happens during triple extraction and a stronger model resolves more of it, but the effect is bounded to that category and does not propagate to the system score. Norm-exception (NI) and temporal-shift (TS), the categories the failure-mode analysis (Section~\ref{sec:failure-modes}) traces to storage and indexing, do not improve ($0.000$ and $-0.030$). Extraction quality is not the binding constraint; the failures the benchmark reports live in how memory is stored and indexed, not in how facts are read from the text.

\section{Per-Category Confidence Intervals}
\label{app:category-cis}

Table~\ref{tab:category-cis} reports a 95\% confidence interval for every condition and category, using the same network bootstrap as Table~\ref{tab:main-results}. MA and DM rest on 22 and 10 questions from 13 and 4 networks because they need a topic the whole group discusses or a member departure (Appendix~\ref{app:data-construction}), so their intervals are wider.

\begin{table*}[t]
  \centering \small
  \setlength{\tabcolsep}{2.5pt}
  \begin{tabular}{l ccccccccc}
    \toprule
    & \multicolumn{9}{c}{Mean score, with 95\% confidence interval below} \\
    \cmidrule(l){2-10}
    \textbf{Condition} & \textbf{SR} & \textbf{GD} & \textbf{MA} & \textbf{AP} & \textbf{TM} & \textbf{NI} & \textbf{RE} & \textbf{TS} & \textbf{DM} \\
    Questions (networks) & 249 (42) & 81 (38) & 22 (13) & 80 (41) & 92 (43) & 54 (37) & 181 (42) & 262 (41) & 10 (4) \\
    \midrule
    \textsc{llm-mini} (full context) & \begin{tabular}[t]{@{}c@{}}0.34\\[-1pt]{\scriptsize [0.28, 0.40]}\end{tabular} & \begin{tabular}[t]{@{}c@{}}0.66\\[-1pt]{\scriptsize [0.54, 0.77]}\end{tabular} & \begin{tabular}[t]{@{}c@{}}0.12\\[-1pt]{\scriptsize [0.04, 0.23]}\end{tabular} & \begin{tabular}[t]{@{}c@{}}0.75\\[-1pt]{\scriptsize [0.65, 0.84]}\end{tabular} & \begin{tabular}[t]{@{}c@{}}0.31\\[-1pt]{\scriptsize [0.26, 0.37]}\end{tabular} & \begin{tabular}[t]{@{}c@{}}0.59\\[-1pt]{\scriptsize [0.47, 0.72]}\end{tabular} & \begin{tabular}[t]{@{}c@{}}0.23\\[-1pt]{\scriptsize [0.17, 0.29]}\end{tabular} & \begin{tabular}[t]{@{}c@{}}0.28\\[-1pt]{\scriptsize [0.23, 0.34]}\end{tabular} & \begin{tabular}[t]{@{}c@{}}0.38\\[-1pt]{\scriptsize [0.28, 0.53]}\end{tabular} \\[3pt]
    \textsc{uncompressed retrieval} & \begin{tabular}[t]{@{}c@{}}0.34\\[-1pt]{\scriptsize [0.29, 0.37]}\end{tabular} & \begin{tabular}[t]{@{}c@{}}0.64\\[-1pt]{\scriptsize [0.51, 0.75]}\end{tabular} & \begin{tabular}[t]{@{}c@{}}0.12\\[-1pt]{\scriptsize [0.03, 0.26]}\end{tabular} & \begin{tabular}[t]{@{}c@{}}0.82\\[-1pt]{\scriptsize [0.73, 0.90]}\end{tabular} & \begin{tabular}[t]{@{}c@{}}0.21\\[-1pt]{\scriptsize [0.16, 0.28]}\end{tabular} & \begin{tabular}[t]{@{}c@{}}0.44\\[-1pt]{\scriptsize [0.30, 0.60]}\end{tabular} & \begin{tabular}[t]{@{}c@{}}0.24\\[-1pt]{\scriptsize [0.19, 0.29]}\end{tabular} & \begin{tabular}[t]{@{}c@{}}0.24\\[-1pt]{\scriptsize [0.19, 0.28]}\end{tabular} & \begin{tabular}[t]{@{}c@{}}0.34\\[-1pt]{\scriptsize [0.17, 0.45]}\end{tabular} \\[3pt]
    \midrule
    \textsc{graphiti} & \begin{tabular}[t]{@{}c@{}}0.11\\[-1pt]{\scriptsize [0.08, 0.14]}\end{tabular} & \begin{tabular}[t]{@{}c@{}}0.62\\[-1pt]{\scriptsize [0.49, 0.74]}\end{tabular} & \begin{tabular}[t]{@{}c@{}}0.06\\[-1pt]{\scriptsize [0.00, 0.17]}\end{tabular} & \begin{tabular}[t]{@{}c@{}}0.46\\[-1pt]{\scriptsize [0.37, 0.56]}\end{tabular} & \begin{tabular}[t]{@{}c@{}}0.05\\[-1pt]{\scriptsize [0.03, 0.08]}\end{tabular} & \begin{tabular}[t]{@{}c@{}}0.41\\[-1pt]{\scriptsize [0.29, 0.53]}\end{tabular} & \begin{tabular}[t]{@{}c@{}}0.12\\[-1pt]{\scriptsize [0.08, 0.16]}\end{tabular} & \begin{tabular}[t]{@{}c@{}}0.07\\[-1pt]{\scriptsize [0.05, 0.09]}\end{tabular} & \begin{tabular}[t]{@{}c@{}}0.00\\[-1pt]{\scriptsize [0.00, 0.00]}\end{tabular} \\[3pt]
    \textsc{langmem} & \begin{tabular}[t]{@{}c@{}}0.08\\[-1pt]{\scriptsize [0.06, 0.11]}\end{tabular} & \begin{tabular}[t]{@{}c@{}}0.57\\[-1pt]{\scriptsize [0.42, 0.70]}\end{tabular} & \begin{tabular}[t]{@{}c@{}}0.15\\[-1pt]{\scriptsize [0.05, 0.29]}\end{tabular} & \begin{tabular}[t]{@{}c@{}}0.39\\[-1pt]{\scriptsize [0.29, 0.48]}\end{tabular} & \begin{tabular}[t]{@{}c@{}}0.06\\[-1pt]{\scriptsize [0.03, 0.08]}\end{tabular} & \begin{tabular}[t]{@{}c@{}}0.44\\[-1pt]{\scriptsize [0.31, 0.57]}\end{tabular} & \begin{tabular}[t]{@{}c@{}}0.06\\[-1pt]{\scriptsize [0.04, 0.09]}\end{tabular} & \begin{tabular}[t]{@{}c@{}}0.09\\[-1pt]{\scriptsize [0.07, 0.12]}\end{tabular} & \begin{tabular}[t]{@{}c@{}}0.15\\[-1pt]{\scriptsize [0.07, 0.21]}\end{tabular} \\[3pt]
    \textsc{mem0} & \begin{tabular}[t]{@{}c@{}}0.11\\[-1pt]{\scriptsize [0.09, 0.15]}\end{tabular} & \begin{tabular}[t]{@{}c@{}}0.58\\[-1pt]{\scriptsize [0.43, 0.72]}\end{tabular} & \begin{tabular}[t]{@{}c@{}}0.05\\[-1pt]{\scriptsize [0.00, 0.11]}\end{tabular} & \begin{tabular}[t]{@{}c@{}}0.31\\[-1pt]{\scriptsize [0.22, 0.41]}\end{tabular} & \begin{tabular}[t]{@{}c@{}}0.05\\[-1pt]{\scriptsize [0.02, 0.08]}\end{tabular} & \begin{tabular}[t]{@{}c@{}}0.24\\[-1pt]{\scriptsize [0.14, 0.35]}\end{tabular} & \begin{tabular}[t]{@{}c@{}}0.06\\[-1pt]{\scriptsize [0.04, 0.09]}\end{tabular} & \begin{tabular}[t]{@{}c@{}}0.07\\[-1pt]{\scriptsize [0.05, 0.09]}\end{tabular} & \begin{tabular}[t]{@{}c@{}}0.10\\[-1pt]{\scriptsize [0.00, 0.18]}\end{tabular} \\[3pt]
    \textsc{cognee} & \begin{tabular}[t]{@{}c@{}}0.09\\[-1pt]{\scriptsize [0.07, 0.12]}\end{tabular} & \begin{tabular}[t]{@{}c@{}}0.58\\[-1pt]{\scriptsize [0.42, 0.72]}\end{tabular} & \begin{tabular}[t]{@{}c@{}}0.02\\[-1pt]{\scriptsize [0.00, 0.07]}\end{tabular} & \begin{tabular}[t]{@{}c@{}}0.25\\[-1pt]{\scriptsize [0.14, 0.37]}\end{tabular} & \begin{tabular}[t]{@{}c@{}}0.10\\[-1pt]{\scriptsize [0.06, 0.14]}\end{tabular} & \begin{tabular}[t]{@{}c@{}}0.20\\[-1pt]{\scriptsize [0.10, 0.31]}\end{tabular} & \begin{tabular}[t]{@{}c@{}}0.05\\[-1pt]{\scriptsize [0.04, 0.07]}\end{tabular} & \begin{tabular}[t]{@{}c@{}}0.01\\[-1pt]{\scriptsize [0.00, 0.02]}\end{tabular} & \begin{tabular}[t]{@{}c@{}}0.07\\[-1pt]{\scriptsize [0.00, 0.12]}\end{tabular} \\[3pt]
    \midrule
    \textsc{subject-mem} & \begin{tabular}[t]{@{}c@{}}0.31\\[-1pt]{\scriptsize [0.26, 0.35]}\end{tabular} & \begin{tabular}[t]{@{}c@{}}0.62\\[-1pt]{\scriptsize [0.50, 0.73]}\end{tabular} & \begin{tabular}[t]{@{}c@{}}0.06\\[-1pt]{\scriptsize [0.01, 0.11]}\end{tabular} & \begin{tabular}[t]{@{}c@{}}0.78\\[-1pt]{\scriptsize [0.68, 0.86]}\end{tabular} & \begin{tabular}[t]{@{}c@{}}0.24\\[-1pt]{\scriptsize [0.19, 0.29]}\end{tabular} & \begin{tabular}[t]{@{}c@{}}0.46\\[-1pt]{\scriptsize [0.33, 0.59]}\end{tabular} & \begin{tabular}[t]{@{}c@{}}0.19\\[-1pt]{\scriptsize [0.16, 0.22]}\end{tabular} & \begin{tabular}[t]{@{}c@{}}0.22\\[-1pt]{\scriptsize [0.18, 0.26]}\end{tabular} & \begin{tabular}[t]{@{}c@{}}0.35\\[-1pt]{\scriptsize [0.25, 0.44]}\end{tabular} \\[3pt]
    \textsc{smg} & \begin{tabular}[t]{@{}c@{}}0.16\\[-1pt]{\scriptsize [0.14, 0.19]}\end{tabular} & \begin{tabular}[t]{@{}c@{}}0.70\\[-1pt]{\scriptsize [0.58, 0.80]}\end{tabular} & \begin{tabular}[t]{@{}c@{}}0.08\\[-1pt]{\scriptsize [0.02, 0.17]}\end{tabular} & \begin{tabular}[t]{@{}c@{}}0.44\\[-1pt]{\scriptsize [0.33, 0.54]}\end{tabular} & \begin{tabular}[t]{@{}c@{}}0.11\\[-1pt]{\scriptsize [0.08, 0.15]}\end{tabular} & \begin{tabular}[t]{@{}c@{}}0.46\\[-1pt]{\scriptsize [0.33, 0.59]}\end{tabular} & \begin{tabular}[t]{@{}c@{}}0.12\\[-1pt]{\scriptsize [0.09, 0.15]}\end{tabular} & \begin{tabular}[t]{@{}c@{}}0.10\\[-1pt]{\scriptsize [0.08, 0.13]}\end{tabular} & \begin{tabular}[t]{@{}c@{}}0.23\\[-1pt]{\scriptsize [0.07, 0.33]}\end{tabular} \\[3pt]
    \bottomrule
  \end{tabular}
  \caption{Per-category scores for every condition on the 43 evaluated networks; the smaller line under each mean is its 95\% confidence interval, and the row under the category codes gives the scored questions and contributing networks per category. Intervals come from 10{,}000 bootstrap resamples over networks; each replicate pools the questions of the resampled networks, matching the question-weighted category means in Table~\ref{tab:main-results}.}
  \label{tab:category-cis}
\end{table*}

\section{Performance by Network Size}
\label{app:tiers}

\begin{figure*}[t!]
  \centering
  \includegraphics[width=0.85\linewidth]{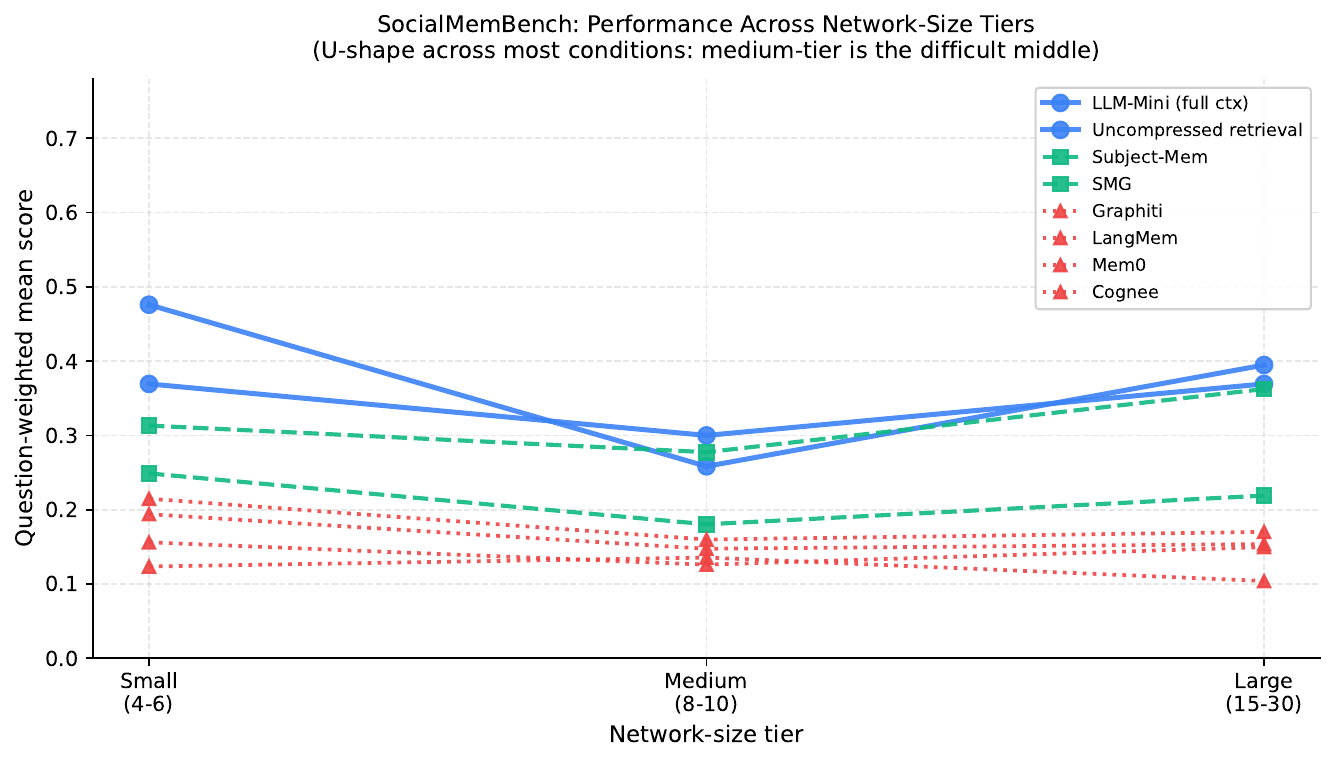}
  \caption{Performance across network-size tiers (visualisation of Table~\ref{tab:tier-results}). Most conditions follow a U-shape with medium-tier as the difficult middle: small-tier corpora concentrate signal per persona, large-tier corpora compensate with more sessions and turns providing supporting attribution evidence. Memory frameworks plateau at 0.10--0.17 across medium and large; the uncompressed retrieval baseline and research probes recover meaningfully at large scale.}
  \label{fig:tier-curve}
\end{figure*}

Performance follows a U-shape across tiers (Figure~\ref{fig:tier-curve}): small-tier corpora concentrate signal per persona, medium-tier sits in a difficult middle, and large-tier recovers as more sessions provide supporting attribution evidence. The pattern is most pronounced for full-context models (\textsc{llm-mini}: 0.476 small, 0.258 medium, 0.395 large). Question-type mix is similar across tiers (AP 7--10\%, TS 22--28\%, SR 19--29\%), and reweighting each tier to the overall mix leaves the medium-tier dip in place, so the dip is not a selection artifact.

Memory frameworks show the same shape at lower amplitude (0.12--0.22 small, 0.13--0.16 medium, 0.10--0.17 large). The plateau is consistent with a fixed retrieval budget: top-$k$ chunks from a growing store return a lower fraction of relevant context per query, and the probability that a fact is stored under the wrong persona increases as more speakers share the same index. Graphiti's large-tier profile shows TS at a 0.078 floor, with NI at 0.263 and RE at 0.124, consistent with merging cascades that depend on persona similarity rather than group size alone.

\begin{figure*}[t!]
  \centering
  \includegraphics[width=0.85\linewidth]{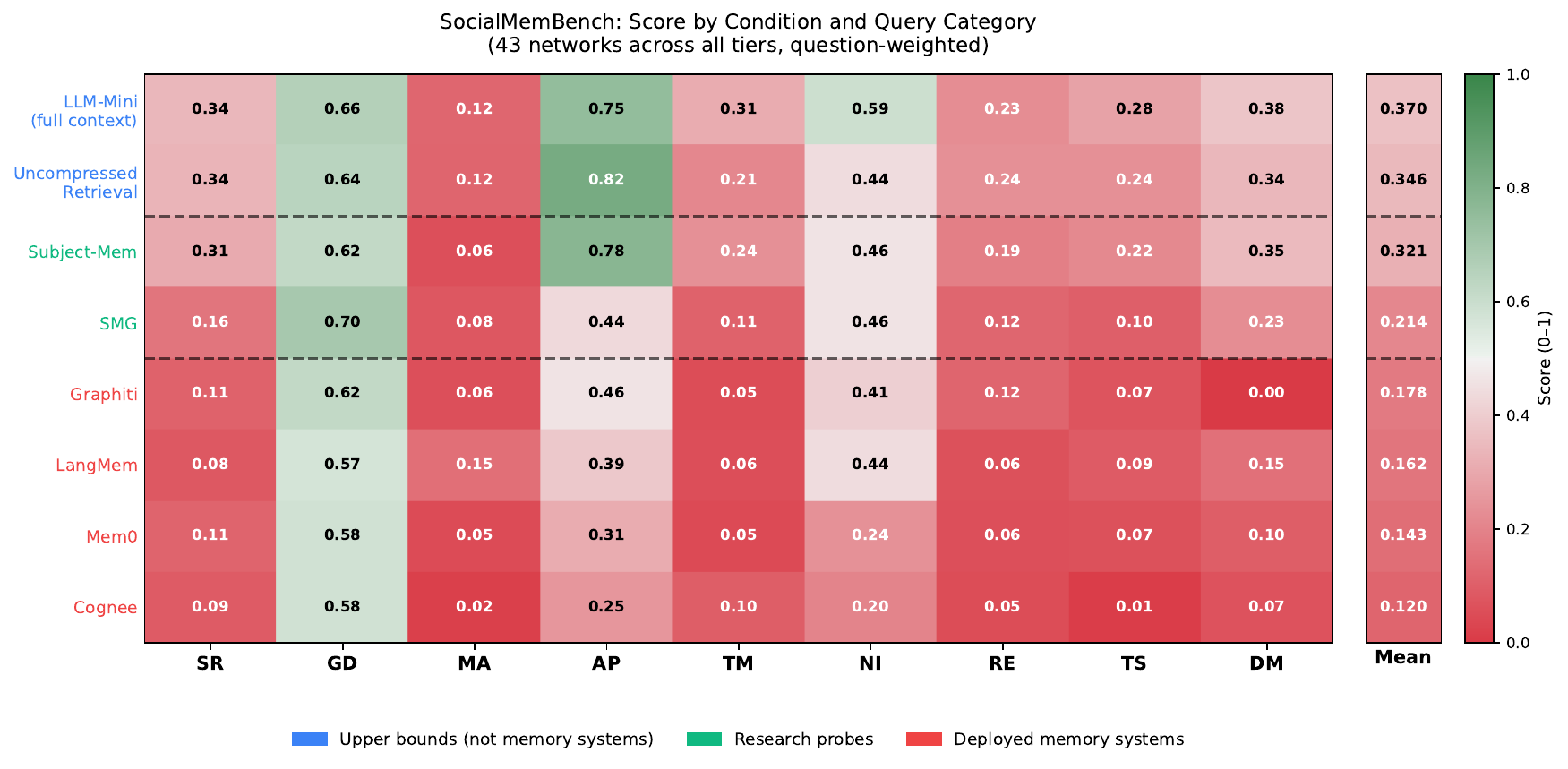}
  \caption{Score by condition and query category (43 networks, question-weighted). Rows are ordered from full-context references (blue) through research probes (green) to memory frameworks (red). The rightmost axis shows each condition's mean score. SMG leads on group decisions (GD: 0.70) and subject-mem leads on attribution probes (AP: 0.78); TM and TS scores are near floor across all four memory frameworks; on DM (departed-member) Graphiti scores 0.00 while Subject-Mem retains 0.35.}
  \label{fig:heatmap}
\end{figure*}
\begin{table*}[t!]
  \centering
  \small
  \begin{tabular}{lccc}
    \toprule
    \textbf{Condition} & \textbf{Small} & \textbf{Medium} & \textbf{Large} \\
    \midrule
    \multicolumn{4}{l}{\textit{Reference conditions}} \\
    \textsc{llm-mini} & 0.476 & 0.258 & 0.395 \\
    \textsc{uncompressed ret.} & 0.369 & 0.300 & 0.369 \\
    \midrule
    \multicolumn{4}{l}{\textit{Memory frameworks}} \\
    \textsc{graphiti} & 0.215 & 0.160 & 0.170 \\
    \textsc{langmem} & 0.194 & 0.147 & 0.154 \\
    \textsc{mem0} & 0.156 & 0.126 & 0.150 \\
    \textsc{cognee} & 0.123 & 0.136 & 0.104 \\
    \midrule
    \multicolumn{4}{l}{\textit{Research probes}} \\
    \textsc{subject-mem} & 0.313 & 0.277 & 0.362 \\
    \textsc{smg} & 0.249 & 0.180 & 0.219 \\
    \bottomrule
  \end{tabular}
  \caption{Question-weighted mean score by network-size tier. Tier definitions and question counts are given in Table~\ref{tab:dataset-stats}; a visualisation of the U-shape across tiers is in Figure~\ref{fig:tier-curve}.}
  \label{tab:tier-results}
\end{table*}
\section{Failure Mode Summary}
\label{app:failure-summary}

\begin{table*}[t!]
  \centering
  \small
  \begin{tabular}{p{3.6cm} p{3.8cm} p{2.4cm} p{3.2cm}}
    \toprule
    \textbf{Failure pattern} & \textbf{Affected adapters} & \textbf{Categories} & \textbf{Fix requires} \\
    \midrule
    Single-stream conflation & Cognee, LangMem, Mem0 & SR, AP & Per-persona attribution slots \\
    Temporal state overwrite & Graphiti, SMG & TS & Ordered fact history (\texttt{EVOLVED\_FROM}) \\
    Entity merging at scale & Graphiti & DM, large-tier NI/RE/TS & Entity-lifecycle preservation \\
    No cross-persona knowledge & All four frameworks & TM & Cross-persona edges (\texttt{KNOWS\_ABOUT}) \\
    Norm-individual conflation & Cognee, LangMem, Mem0, Graphiti & NI, GD & Dissent-aware schema \\
    \bottomrule
  \end{tabular}
  \caption{Five failure patterns observed in evaluated adapter configurations, the systems affected, and the categories where they surface. Claims are specific to the adapter versions and configurations tested. No evaluated system addresses all five. Categories give the measurement target for each failure mode.}
  \label{tab:failure-modes}
\end{table*}
\section{Pipeline Overview}
\label{app:pipeline}

\begin{figure*}[t!]
  \centering
  \includegraphics[width=\linewidth]{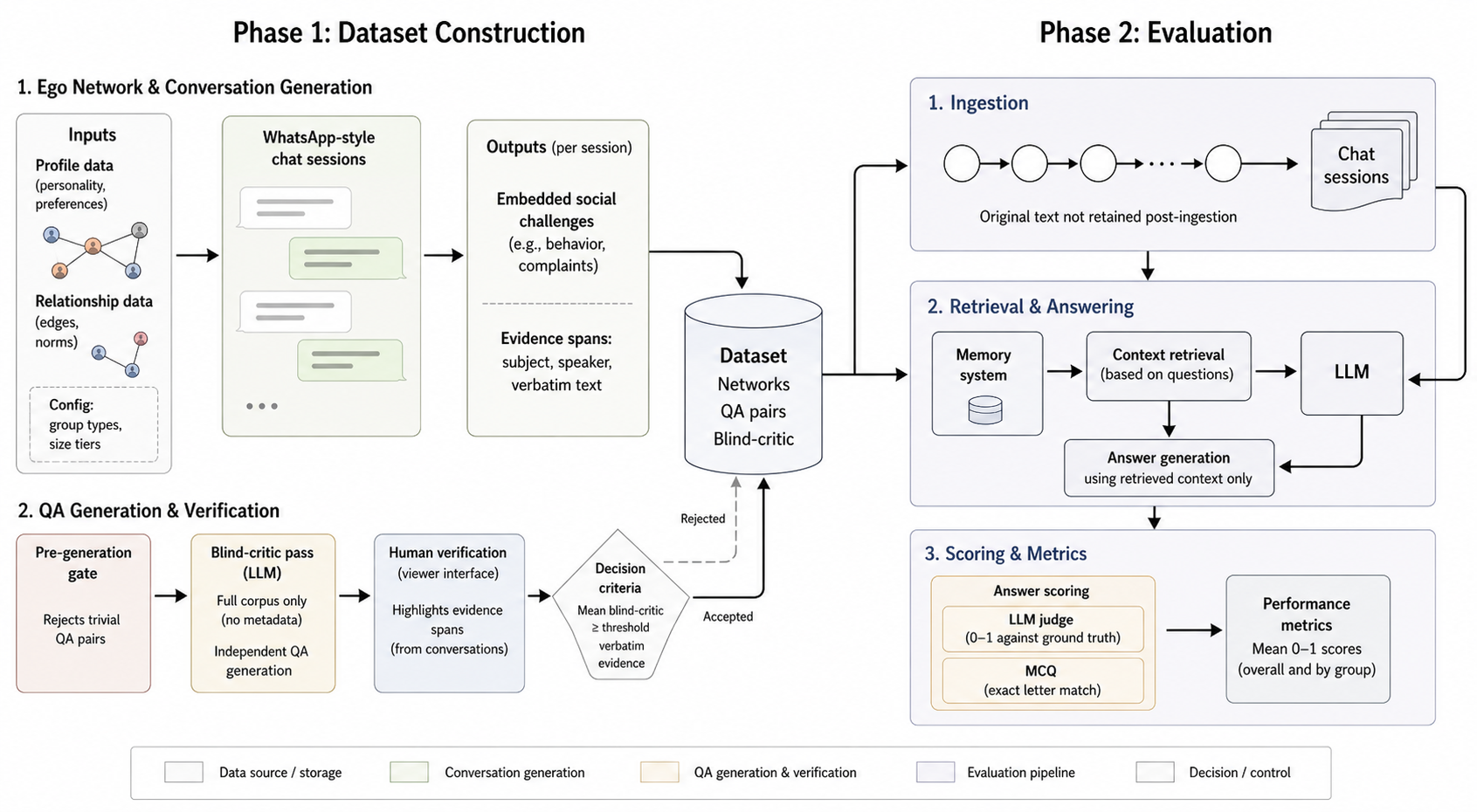}
  \caption{SocialMemBench pipeline. Phase 1 (left) covers dataset construction: persona network and conversation generation with implicitly planted social challenges, followed by QA generation with a pre-generation gate, blind-critic pass, and human verification via the data audit viewer. Phase 2 (right) covers evaluation: session-by-session ingestion into a memory system, retrieval and answering using only retained context, and LLM judge scoring.}
  \label{fig:pipeline}
\end{figure*}

Figure~\ref{fig:pipeline} diagrams the end-to-end SocialMemBench workflow. The left side covers data construction, executed once per network: ego-network generation, conversation generation with implicitly planted social challenges across $S$ sessions, then QA generation with the pre-generation inference gate (Section~\ref{sec:data-gen}), the blind-critic acceptance pass, and human review through the audit viewer. A network only enters the released dataset after all three gates pass. The right side covers evaluation: each condition ingests one session at a time into its memory system, the system's retrieve interface serves a held-out QA pair, and the answerer produces a response that an LLM judge scores against the ground-truth answer. Memory state persists across sessions within a network and is reset between networks, isolating per-network behaviour.

\section{Data Audit Viewer}
\label{app:viewer}

The repository includes a browser-based data audit viewer used during dataset construction for human review. It runs locally against the released data files and requires no API key.

\begin{verbatim}
cd viewer && python server.py
# Open http://localhost:8080
\end{verbatim}

The viewer has three tabs. The \textbf{Chat} tab renders each network's conversation sessions as a WhatsApp-style thread, with planted-challenge turns highlighted and evidence spans underlined inline; clicking a highlighted turn shows the challenge type, subject persona, and verbatim evidence span. The \textbf{Graph} tab shows the persona network as an interactive D3 force graph, with nodes coloured by persona and edges labelled by relationship type, closeness, and sentiment. The \textbf{QA} tab lists all QA pairs for the network with their evidence anchors, query type, difficulty label, and QC flags; pairs can be filtered by query type and searched by question text.

During dataset construction, all 1,031 QA pairs were reviewed in this interface to check plausibility, correct evidence attribution, and answer completeness. Human review is independent of the automated blind critic: the viewer presents the conversation and the gold QA (question, answer, and evidence anchors) but not the blind critic's generated answers or recoverability scores, so reviewers do not anchor on the automated signal. The blind critic in turn never sees the human judgments, so the two checks give independent evidence of answer quality. The viewer is released as part of the benchmark so that reviewers and future users can inspect any network and verify ground-truth anchors without writing code.

\begin{figure*}[t!]
  \centering
  \includegraphics[width=\linewidth]{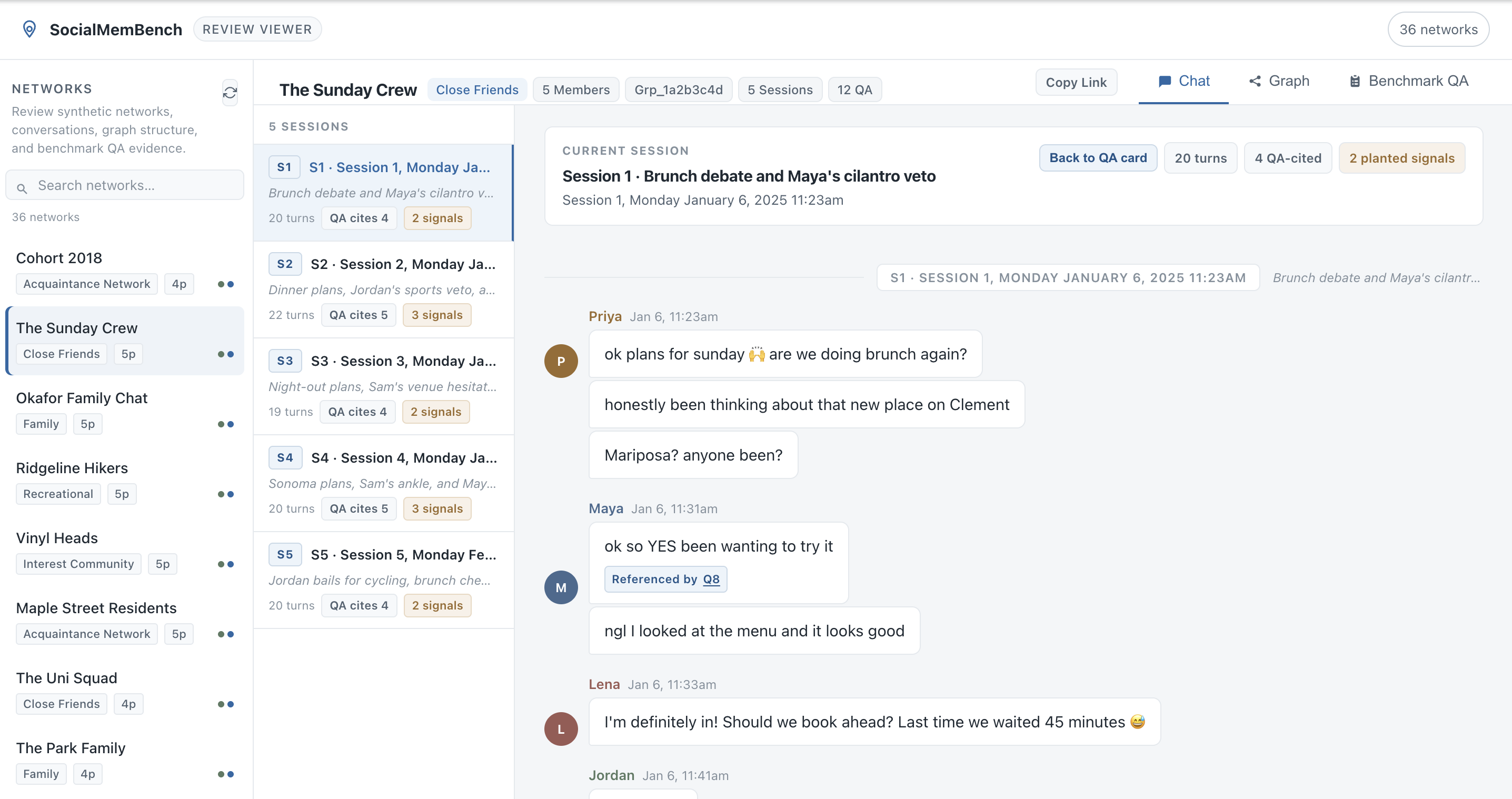}
  \caption{Data audit viewer showing the Chat tab. Turns containing planted challenges are highlighted; clicking a highlighted turn opens the challenge panel (right) with the challenge type, subject persona, and verbatim evidence span. The QA tab (not shown) links each question to its evidence anchors in the same interface.}
  \label{fig:viewer}
\end{figure*}
\section{Representative Prompts}
\label{app:prompts}

The following prompts are reproduced verbatim from \texttt{eval/config/prompts.yaml} and \texttt{eval/adapters/subject\_rag\_adapter.py} (line breaks added to fit the column). Template variables (\texttt{\{context\}}, \texttt{\{question\}}, etc.) are filled at runtime.

\subsection{Open-Ended Answerer (Memory System Conditions)}

This prompt is used when a memory system's retrieved output is passed to GPT-4o-mini as context. It instructs the model to use only what was retrieved and provides per-category answering guidance.

\begin{verbatim}
You are evaluating a social group AI memory
  system.
You have retrieved memories from a group chat
  and must answer
an open-ended question. Answer ONLY using the
  retrieved
memories. Use no outside knowledge or guesses.

Pay attention to WHO said what and WHEN.
Attribution and timing matter.

Answering guidance by question type:
- Single preference (Q1): Name the person and
  state their
  preference. Be specific.
- Group decision (Q2): State the decision AND
  name any
  dissenter(s); implicit dissent counts.
- Multi-contact (Q3): Cover every group
  member; say "unclear"
  for those with no signal.
- Theory of mind (Q5): State BOTH the
  preference AND who
  revealed it, with the specific observable
    action that shows
  they already knew.
- Temporal shift (Q8): State the OLD
  preference, the NEW
  preference, and what triggered the change.

[MEMORIES]
{context}

[QUESTION]
{question}
\end{verbatim}

\subsection{LLM Judge}

The judge receives the question, the gold answer, and the generated answer and returns a JSON score. Three critical rules apply before the general rubric: brevity is not penalised, wrong attribution caps the score at 0.3, and a refusal to answer scores 0.0.

\begin{verbatim}
Score the generated answer against the gold
  answer
on a 0.0–1.0 scale.

CRITICAL RULES:
1. A concise correct answer scores THE SAME as
  a verbose one.
2. Attributing a fact to the wrong person ->
  score <= 0.3.
3. "NOT ENOUGH INFORMATION" always scores 0.0.

General rubric:
- 1.0: Core fact correct, correct attribution,
  all parts present
- 0.7–0.9: Right answer but one minor part
  missing
- 0.4–0.6: Right direction but wrong detail or
  missing element
- 0.0–0.3: Wrong attribution, wrong fact, or
  no answer

Q-type rules:
- Q3: score = fraction of group members
  correctly recalled
- Q4: correct speaker + foil not named → 1.0;
  foil also
  named → 0.7; wrong speaker → 0.0–0.3
- Q5: both preference AND observable evidence
  of prior
  knowledge required for 1.0; one part → 0.5
- Q8: old state (+0.33) + new state (+0.33) +
  trigger (+0.33) = 1.0

Question: {question}
Gold answer: {golden_answer}
Generated answer: {generated_answer}

Output JSON only:
{"score": 0.0, "rationale": "one sentence"}
\end{verbatim}

\subsection{Subject-Indexed Fact Extraction (Subject-Mem)}

At ingestion time, Subject-Mem calls GPT-4o-mini once per turn to extract (subject, fact) pairs, where the subject is the person a fact is \emph{about}, not the person who said it. This single change to the indexing step is the only difference from the uncompressed retrieval baseline.

\begin{verbatim}
You are reading a message from a group chat.
  Extract any
facts about specific people mentioned.

Message: [{speaker}]: {message}

For each fact, output a JSON object:
  "subject": the person the fact is about
  "fact": a concise statement of the fact

Return a JSON array. Return [] if no
  person-specific facts.
Only include facts clearly about a named
  person.

Examples:
Input: [Nadia]: Rafe always goes to the pub
  after long runs
Output: [{"subject": "Rafe",
          "fact": "always goes to the pub
            after long runs"}]

Input: [Tom]: Great session today everyone
Output: []
\end{verbatim}

\end{document}